\documentclass[sigconf, nonacm]{acmart}
\usepackage{amsmath,amsfonts}
\usepackage{graphicx}
\usepackage{textcomp}
\usepackage{xcolor}

\usepackage{amsmath}
\usepackage{paralist}
\usepackage{enumitem}
\usepackage[update,prepend]{epstopdf}
\usepackage{bbding}
\usepackage{tabularx}
\usepackage{multirow}
\usepackage{ifsym}
\usepackage{color}
\usepackage{float}
\usepackage{algpseudocode}
\usepackage{threeparttable}
\usepackage{subfigure}
\usepackage{pdfpages}
\usepackage{graphicx}
\usepackage[inkscapelatex=false]{svg}
\usepackage{hyperref}
\usepackage{amsmath,amsfonts}
\usepackage{weiwAlgorithm}
\usepackage{weiwMath}
\makeatletter

\newcommand{\Rmnum}[1]{\expandafter\@slowromancap\romannumeral #1@}
\makeatother
\usepackage{graphicx}
\usepackage{textcomp}
\usepackage{xcolor}
\usepackage{float}
\usepackage{url}
\usepackage{pifont}

\usepackage{amsmath}

\newcommand{\hospital}{\textsf{Hospital}\xspace}
\newcommand{\flights}{\textsf{Flights}\xspace}
\newcommand{\soccer}{\textsf{Soccer}\xspace}
\newcommand{\beers}{\textsf{Beers}\xspace}
\newcommand{\inpatient}{\textsf{Inpatient}\xspace}
\newcommand{\facilities}{\textsf{Facilities}\xspace}

\newcommand{\bclean}{\textsf{BClean}\xspace}       
\newcommand{\bcleanI}{$\textsf{BClean}$\xspace}    
\newcommand{\bcleanII}{$\textsf{BClean}_\textsf{PI}$\xspace}    
\newcommand{\bcleanIII}{$\textsf{BClean}_\textsf{PIP}$\xspace}    
\newcommand{\bcleannouc}{$\textsf{BClean}_\textsf{-UC}$\xspace}    
\newcommand{\pclean}{\textsf{PClean}\xspace}
\newcommand{\raha}{\textsf{Raha}\xspace}
\newcommand{\baran}{\textsf{Baran}\xspace}
\newcommand{\garf}{\textsf{Garf}\xspace}
\newcommand{\rahabaran}{\textsf{Raha+Baran}\xspace}
\newcommand{\holoclean}{\textsf{HoloClean}\xspace}
\newcommand{\bcleanhill}{\textsf{BClean+Hill}\xspace}

\newcommand{\wys}[1]{{\color{blue}{#1}}}
\newcommand{\eat}[1]{}
\newcommand{\kw}[1]{{\textsf{#1}}\xspace}

\newcommand{\etitle}[1]{\vspace{1ex}\noindent{\em\underline{#1}}}

\newcommand{\sstab}{\rule{0pt}{8pt}\\[-1.8ex]}

\newcommand{\bi}{\begin{itemize}}
\newcommand{\ei}{\end{itemize}}
        {\end{itemize}} 

\newcommand{\be}{\begin{enumerate}}
\newcommand{\ee}{\end{enumerate}}
\newcommand{\beqn}{\begin{eqnarray}}
\newcommand{\eeqn}{\end{eqnarray}}

\newcounter{ccc}

\newcounter{example}
\renewcommand{\theexample}{\arabic{example}}
\newenvironment{example}{\begin{em}
        \vspace{1.5ex}
        \refstepcounter{example}
        {\noindent\bf Example \theexample:}}{
  \end{em} 
  \vspace{1.5ex}}

\renewcommand{\texttt}[1]{{\small\textsf{#1}}}

\definecolor{gray}{rgb}{0.5,0.5,0.5}

\newlist{myitemize}{itemize}{3}
\setlist[myitemize,1]{label=$\circ$,leftmargin=3.6ex}
\setlist[myitemize,2]{label=$\bullet$,leftmargin=4.0ex}
\setlist[myitemize,3]{label=$\diamond$, leftmargin=3.5ex}

        \newcommand{\mei}{\end{myitemize}\vspace{0.6ex}}

\makeatletter
\newcommand\figcaption{\def\@captype{figure}\caption}
\newcommand\tabcaption{\def\@captype{table}\caption}
\makeatother

\newcommand{\flight}{\kw{Flight}}

\author{Jianbin Qin$^{1,2}$, Sifan Huang$^1$, Yaoshu Wang$^2$, Jing Zhu$^1$, Yifan Zhang$^1$,} 
\author{Yukai Miao$^3$, Rui Mao$^{1,2}$, Makoto Onizuka$^4$, Chuan Xiao$^{4,5}$}
\affiliation{
  \institution{$^1$Shenzhen University \hspace{8ex}
    $^2$Shenzhen Institute of Computing Sciences \hspace{8ex}
    $^3$Zhongguancun Laboratory 
  }
  \institution{$^4$Osaka Univeristy \hspace{8ex}
    $^5$Nagoya University
  }
  \country{}
}
\email{{qinjianbin, mao}@szu.edu.cn, {huangsifan2020, 2300271063, zhangyifan2021}@email.szu.edu.cn}
\email{yaoshsuw@sics.ac.cn, miaoyk@zgclab.edu.cn, {onizuka, chuanx}@ist.osaka-u.ac.jp}

\begin{document}

\title{BClean: A Bayesian Data Cleaning System}

\begin{abstract}
  There is a considerable body of work on data cleaning which employs various 
  principles to rectify erroneous data and transform a dirty dataset into a 
  cleaner one. One of prevalent approaches is probabilistic methods, including 
  Bayesian methods. However, existing probabilistic methods often assume a 
  simplistic distribution (e.g., Gaussian distribution), which is frequently 
  underfitted in practice, or they necessitate experts to provide a complex 
  prior distribution (e.g., via a programming language). This requirement is 
  both labor-intensive and costly, rendering these methods less suitable for 
  real-world applications. In this paper, we propose \bclean, a 
  \textbf{B}ayesian \textbf{C}leaning system that features automatic Bayesian 
  network construction and user interaction. We recast the data cleaning problem 
  as a Bayesian inference that fully exploits the relationships between 
  attributes in the observed dataset and any prior information provided by 
  users. To this end, we present an automatic Bayesian network construction 
  method that extends a structure learning-based functional dependency discovery 
  method with similarity functions to capture the relationships between 
  attributes. Furthermore, our system allows users to modify the generated 
  Bayesian network in order to specify prior information or correct inaccuracies 
  identified by the automatic generation process. We also design an effective 
  scoring model (called the compensatory scoring model) necessary for the 
  Bayesian inference. To enhance the efficiency of data cleaning, we propose 
  several approximation strategies for the Bayesian inference, including graph 
  partitioning, domain pruning, and pre-detection. By evaluating on both 
  real-world and synthetic datasets, we demonstrate that \bclean is capable of 
  achieving an F-measure of up to 0.9 in data cleaning, outperforming existing 
  Bayesian methods by 2\% and other data cleaning methods by 15\%. Our source code is available at \url{https://github.com/yyssl88/BClean} .
\end{abstract}

\maketitle

\section{Introduction}
\label{sec:intro}
Data cleaning is an essential step in preprocessing data for downstream 
applications such as data analysis and machine learning (ML) model construction. 
Existing data cleaning solutions utilize user-defined rules, outlier detection 
techniques, crowdsourcing, or knowledge bases to identify errors in the data~\cite{heidari2019holodetect,abedjan2016detecting,mahdavi2019raha}, 
and subsequently carry out data repairing~\cite{DBLP:conf/aistats/LewASM21,DBLP:journals/pvldb/RekatsinasCIR17,DBLP:journals/pvldb/MahdaviA20}. The majority of research on data cleaning has been dedicated to error 
correction using discriminative models based on integrity constraints~\cite{DBLP:journals/pvldb/YakoutENOI11,DBLP:conf/icdt/KolahiL09,DBLP:conf/sigmod/DallachiesaEEEIOT13}, 
external data~\cite{fan2014interaction,DBLP:conf/sigmod/ChuMIOP0Y15,DBLP:conf/sigmod/ZhengWLCF15,DBLP:conf/sigmod/WangT14,rezig2019towards}, statistical 
approaches~\cite{DBLP:journals/corr/abs-2011-04730, mayfield2009statistical}, 
ML techniques~\cite{DBLP:conf/sigmod/YakoutBE13}, or hybrid methods~\cite{DBLP:conf/icdt/BertossiKL11,DBLP:journals/pvldb/MahdaviA20}. 
Despite these approaches, certain challenges persist. For instance, 
some solutions based on integrity constraints often encounter 
difficulties in real-world applications, as they require expert input 
for rule creation. Solutions that rely on external data incur 
substantial costs associated with expert information collection. 
Moreover, ML methods face potential confusion of learning models when 
generating feature vectors from noisy data.

To address these shortcomings, probabilistic 
methods~\cite{DBLP:conf/icdm/KubicaM03,li2019statistical} 
have been proposed for data cleaning in a data-driven manner. These 
methods employ probabilistic inference~\cite{DBLP:journals/ker/Parsons11a} 
for data cleaning, such as using probabilistic graphical models 
(PGM). Furthermore, probabilistic programming languages 
(PPL)~\cite{bingham2019pyro, carpenter2017stan, milch20071} have been 
developed as powerful tools for describing and executing probabilistic 
models. Another typical method is HoloClean~\cite{DBLP:journals/pvldb/RekatsinasCIR17}, 
in which probabilistic inference acts as a feature generator to 
generate the degrees of validity of denial constraints (DCs), one type of 
dependency rules. Configuring rules in HoloClean is simpler than in PPL. Despite 
achieving robust cleaning performance in a few empirical evaluations, HoloClean 
is a semi-supervised method that requires manually annotated data. Other 
semi-supervised methods, such as \raha~\cite{mahdavi2019raha} and 
\baran~\cite{DBLP:journals/pvldb/MahdaviA20}, feature automatic rule discovery. 
They need labels for around 20 tuples to perform error detection and correction. 
However, error propagation may occur from detection to correction, despite ease 
of deployment. Currently, probabilistic inference-based data cleaning remains an 
active area of study, as PGMs offer natural advantages when modeling 
dirty data as generative models~\cite{DBLP:journals/corr/abs-1204-3677}.

Recently, Bayesian methods, a subset of probabilistic methods, have been explored 
and have demonstrated promising results~\cite{DBLP:conf/aistats/LewASM21}. For 
instance, existing 
studies~\cite{DBLP:journals/jdiq/DeHMCK16,DBLP:journals/corr/abs-1204-3677,DBLP:journals/pvldb/ZhaoRGH12} 
employ Bayesian networks to correct errors. These methods encompass Bayesian 
network construction and inference. The network construction involves structure 
learning and parameter estimation from the observed data, whereas the inference 
aims to determine the most probable value to fill in each data cell, given other 
attributes of the same tuple and the learned data distribution. If the inferred 
value differs from the original value in the cell, the inferred value replaces the 
original one, serving as a correction for the cell.

\begin{table*}[!t]
  \small
  \caption{Customer table.}  
  \vspace{-2ex}
  \centering
  \begin{tabular}{ccccccccc}  
  \hline  
  \textbf{Tid}  & \textbf{Name}  & \textbf{Department} & \textbf{Jobid} & \textbf{City} & \textbf{State}  & \textbf{ZipCode} & \textbf{InsuranceCode} & \textbf{InsuranceType} \\  
  \hline  
  1 & Johnny.R & 315 w hickory st & 25676000 & sylacauga & CA & 35150 & 2567600035150 &  \\ 
  
  2 & Johnny.R & \textcolor{red}{400 northwood dr} & \textcolor{red}{25676x00} & sylacauga & \textcolor{red}{KT} & 35150 & 2567600035150 & Normal \\  
  
  3 & Johnny.R & \textcolor{red}{315 w hicky st} & 25676000 & sylacauga & CA & 35150 & 2567600035150 & Normal \\
  
 4 & Henry.P & 400 northwood dr & 25600180 & centre & KT &  & 2560018035960 & Low \\ 
  
  5 & Henry.P & \textcolor{red}{400 nprthwood dr} & 25600180 & centre & \textcolor{red}{NY} & \textcolor{red}{3960} & \textcolor{red}{25600v5960} & \textcolor{red}{High} \\
  
  6 & Henry.P &  & 25600180 & centre & KT & 35960 &  & Low \\ 
  \hline  
  \end{tabular}
  \label{tab:ExampleDataset}
\end{table*}

\noindent\textbf{Motivation.} 
Existing Bayesian methods face three key challenges: 
\begin{inparaenum} [(1)]
    \item a high dependence on accurate domain knowledge from experts, which 
    is required for PPL~\cite{DBLP:conf/aistats/LewASM21}, and necessitates 
    user interaction and domain knowledge editing,
    \item inefficiency of inference using Bayesian 
    networks~\cite{DBLP:journals/jdiq/DeHMCK16,DBLP:journals/corr/abs-1204-3677}, 
    in which variable elimination following the topological order of the nodes 
    incurs significant computational cost, and 
    \item the presence of errors in the constructed Bayesian network originating 
    from noisy data, which may propagate into subsequent inferences, resulting in 
    inferior inference performance.
\end{inparaenum}

\begin{example}
  Table~\ref{tab:ExampleDataset} shows a Customer table that includes 6 tuples 
  with 9 attributes. Most existing methods can accurately identify and correct 
  errors in Tuples 1 -- 3, given that the context is relatively clean and complete. 
  Functional dependency (FD) learning~\cite{DBLP:conf/sigmod/ZhangGR20} can be 
  employed to discover the FD $\kw{InsuranceCode} \rightarrow \kw{InsuranceType}$ 
  to correct the missing value ``$\texttt{NULL} \rightarrow \texttt{Normal}$'' in 
  the first line, and another FD $\kw{ZipCode} \rightarrow \kw{State}$ to correct 
  the error ``$\texttt{KT} \rightarrow \texttt{CA}$'' in the second line. Repairs 
  such as ``$\texttt{25676x00} \rightarrow \texttt{25676000}$'' and 
  ``$\texttt{315 w hicky st} \rightarrow \texttt{315 w hickory st}$'' can be 
  identified using the context~\cite{DBLP:journals/pvldb/MahdaviA20}. In contrast, 
  Tuples 4 -- 6 in Table~\ref{tab:ExampleDataset} present more of a challenge due 
  to the presence of numerous critical errors. The values for $\kw{Department}$, 
  $\kw{InsuranceCode}$, $\kw{InsuranceType}$, and $\kw{State}$ cannot be 
  definitively inferred due to little evidence from $\kw{Name}$, $\kw{Jobid}$, and 
  $\kw{ZipCode}$. Due to the lack of observations, the errors ``\texttt{400 
  nprthwood dr}'', ``\texttt{NY}'', ``\texttt{3960}'', ``\texttt{25600v5960}'', 
  and ``\texttt{High}'' may be perceived as correct values by probabilistic models. 
\end{example}

To address these problems, many existing data cleaning methods incorporate 
external data~\cite{DBLP:conf/sigmod/ChuMIOP0Y15,DBLP:journals/pvldb/RekatsinasCIR17}. 
In Bayesian methods, domain knowledge plays a crucial role in encoding data 
distribution and Bayesian network structure. 
PClean~\cite{DBLP:conf/aistats/LewASM21} is a state-of-the-art Bayesian method, 
whose network construction and the prior knowledge are hand-crafted. This 
requires specification of data types, compliant distributions, and possible 
noises (e.g., uncommon symbols or Gaussian noise). However, authoring the PPL 
code in PClean introduces a steep learning curve and high user cost. For example, 
PClean requires users to precisely partition Table~\ref{tab:ExampleDataset} into 
four parts: $P_1 = \{\kw{Name} \sim \kw{Dist}(\theta_{1})$, $\kw{Name} \sim 
\kw{Dist}(\theta_{2}), \kw{Jobid} \sim \kw{Dist}(\theta_{3}))\}$; 
$P_2 = \{\kw{City} \sim \kw{Dist}(\theta_{4})$, $\kw{State} \sim 
\kw{Dist}(\theta_{5}), \kw{ZipCode} \sim \kw{Dist}(\theta_{6})\}$; 
$P_3 = \{\kw{InsuranceCode} \sim \kw{Dist}(\theta_{7})$, $\kw{InsuranceType} \sim 
\kw{Dist}(\theta_{8})\}$; and $P_4 = \{P_1, P_2, P_3, \kw{error\_dist}(\theta_{9}) \}$, 
where $A \sim \kw{Dist}(\theta)$ in each part denotes an attribute $A$ that 
follows a distribution with parameter $\theta$. It is hard for users to specify 
such partition, and the specification itself is error-prone. 

\noindent\textbf{Our approach.} 
In this paper, we introduce \bclean, an unsupervised Bayesian data 
cleaning system. Given a noisy relational dataset, \bclean performs error 
correction in two stages: the construction stage and the inference stage. In the 
construction stage, \bclean automatically constructs a Bayesian network from the 
dataset and provides a user interaction function that optionally collects 
lightweight domain knowledge from users to fine-tune the Bayesian network. In the 
inference stage, \bclean leverages the dataset and any constraints specified by 
users to develop a compensatory scoring model, which acts as a supplementary model 
providing additional statistical information for inference. Notably, it 
approximates the value of a term in the formula of Bayesian inference, thereby 
addressing the problem of error amplification caused by inaccurate Bayesian network 
construction from dirty data. \bclean processes each cell in the dataset, using the 
constructed Bayesian network and the compensatory scoring model to compute the 
probability of candidate repair options, based on the evidence from other 
attributes of the same tuple. Furthermore, \bclean incorporates a series of 
optimization techniques to boost efficiency, including graph partitioning, domain 
pruning, and pre-detection. An experimental evaluation on real-world and synthetic 
datasets shows that \bclean can achieve an F-measure of up to 0.9 in data cleaning, 
outperforming existing Bayesian methods by 2\% and other data cleaning methods by 
15\%. 

In \bclean, akin to \pclean but more straightforward, users only need to define 
constraints by specifying simple expressions (e.g., categories of values, 
minimum/maximum lengths of attributes, minimum/maximum values of numerical 
attributes, whether null values are allowed, and regular expressions) instead of 
adhering to a specific distribution. As a result, \bclean incurs substantially 
lower user costs than existing solutions (e.g., \pclean), which require users to 
learn PPL to inject prior knowledge. Importantly, the user constraints supported in 
\bclean are not limited to the aforementioned expressions but can be any function 
that returns a binary output, thus covering a wide range of constraint forms such as
dependency rules (FDs and DCs), arithmetic expressions, and even deep neural 
networks. Nevertheless, our experiments show that using simple expressions is 
sufficient to deliver satisfactory data cleaning results. 

Our contributions are summarized as follows:
\begin{itemize}[leftmargin=*]
  \item We propose a Bayesian-inference-based data cleaning system that repairs 
  errors in datasets using a Bayesian network.
  \item We adapt an existing FD discovery method to automatically generate a Bayesian 
  network, and provide users with an interface to optionally modify the Bayesian 
  network to specify prior knowledge of the dataset.
  \item We design a compensatory scoring model for Bayesian inference, leveraging 
  user-specified constraints to filter obviously incorrect values and rank candidate 
  repair options.
  \item We introduce a series of approximation techniques, including graph 
  partitioning, domain pruning, and pre-detection, to enhance the efficiency of data 
  cleaning.
  \item We conduct experiments on real-world and synthetic datasets. The results 
  demonstrate the effectiveness and the efficiency of our system and its 
  competitiveness with alternative solutions, especially those methods based on 
  Bayesian inference.
\end{itemize}

The remainder of this paper is organized as follows. Section~\ref{sec:preliminaries}
introduces preliminaries. Section~\ref{sec:framework} offers an overview of \bclean and its 
data-driven modeling. Sections~\ref{sec:network} -- \ref{sec:optimization} present the 
technical details of \bclean. Section\ref{sec:exp} reports the experimental results. 
Section~\ref{sec:related} reviews related work. Section~\ref{sec:concl} concludes the 
paper.


\section{Preliminaries}
\label{sec:preliminaries}


\begin{figure}[t]
    \centering
    \includegraphics[width=\linewidth]{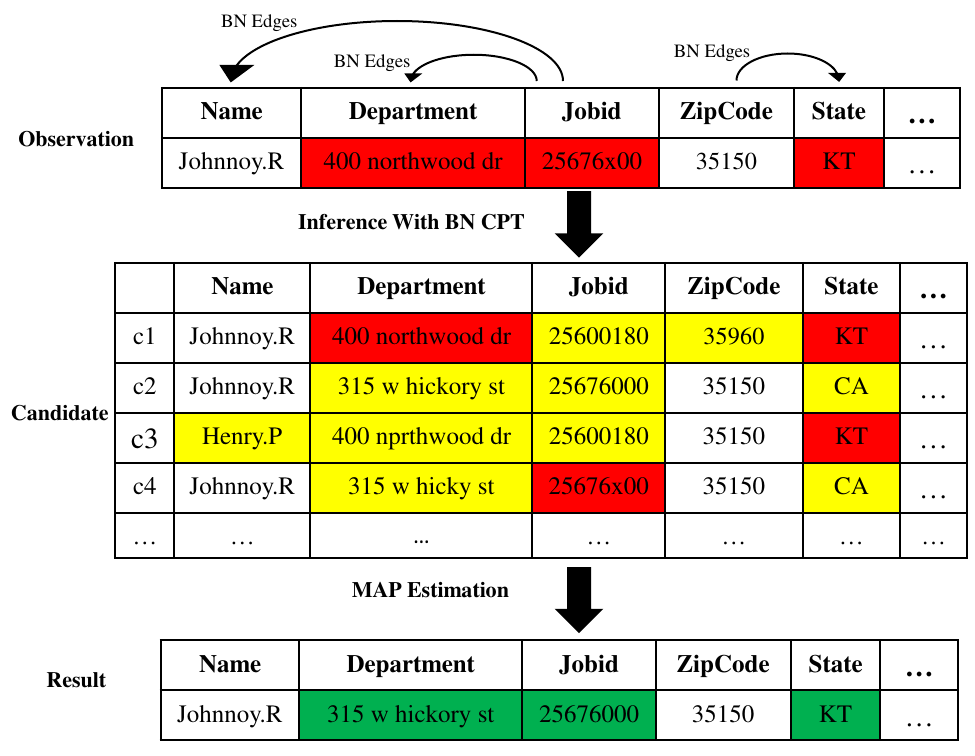}
    \caption{A running example of \bclean. Red, yellow, and green represent 
    erroneous, candidate, and clean cell values.}
    \label{fig:BClean-example}
\end{figure} 

The aim of \bclean is to repair erroneous data in a structured dataset 
$D$ by employing a Bayesian network (BN, a.k.a. Bayes network or belief 
network), thereby producing a cleaner 
version of the dataset. The observed dataset $D$ comprises $n$ tuples, 
$\set{T_{1}, \ldots, T_{n}}$, and is characterized by $m$ attributes: 
$\mathcal{A} = \set{A_{1}, \ldots, A_{m}}$. The domain of attribute 
$A_{j}$ is denoted as $dom(A_{j})$. $T_{i}[A_{j}]$ represents the 
observation of the $j$-th attribute in the $i$-th tuple. To repair the 
dataset, \bclean considers candidate values $c \in dom(A_{j})$ and 
infers the most probable candidate value $c^*$ to replace $T_{i}[A_{j}]$ 
for all $i \in [1, n]$ and $j \in [1, m]$. Moreover, candidate values 
must satisfy the constraints specified by the user; e.g., the ZIP code in 
the US must be five digits. We employ a user constraint (UC) to verify if 
the input (a cell, tuple, or a dataset) adheres to the user-specified 
constraints. A UC is a function $UC(\cdot)$ that returns $1$ if the input 
satisfies the user constraints, or $0$ otherwise. It can be any function 
that returns a binary output, such as rules (e.g., DCs and FDs), 
arithmetic expressions, regular expressions, or even deep neural networks. 
As the interaction should be easy to use, we primarily focus on the 
following UCs in this paper: 
\begin{inparaenum} [(1)]
    \item minimum/maximum attribute lengths (or minimum/maximum values for 
    numerical attributes), 
    \item non-null constraints, and 
    \item simple regular expressions for digits and dates.
\end{inparaenum}

The above UCs are straightforward. The first pertains to attribute 
statistics, the second mandates value assignment, and the third specifies 
the attribute's format. Even for users unfamiliar with regular expressions, 
numerous online tools exist for generating them from examples, such as the 
one detailed in \cite{DBLP:journals/tkde/BartoliLMT16}, with its online 
demonstration available at \cite{regex-generator}. 
The underlying rationale is that these UCs, serving as prior 
knowledge about the dataset, do not necessitate database expertise. 
Therefore, it is unnecessary to presume user familiarity with databases, 
particularly advanced techniques used in \holoclean, including conditional 
FDs~\cite{cfdclean} and metric FDs~\cite{DBLP:conf/icde/KoudasSSV09}.
Additionally, users are not required to label data for many tuples 
(approximately 40, as in \rahabaran). This approach makes our solution more 
broadly applicable. For instance, improving data preprocessing is a 
relatively relevant challenge in economics and social science 
research~\cite{ale2023using}. While model accuracy is crucial in these 
fields, we cannot expect researchers to be versed in concepts like 
functional dependencies, to devote substantial time to acquiring these 
database skills, or to drill down to the dataset to label the correctness 
of certain values in a record-by-record manner.

By inferring the value of each cell in $D$, we obtain the cleaned dataset 
$D^*$. Figure~\ref{fig:BClean-example} depicts an example of applying \bclean to 
a dataset. Although a dataset typically contains many tuples, we 
illustrate only one tuple in this example. The nodes in the BN represent 
atributes. The edges in the BN represent the dependency relationships between 
attributes. Candidate values are generated for each cell, and their 
probabilities are computed by maximum a posteriori (MAP) estimation using the BN.

\begin{figure*}[t]
    \centering
    \includegraphics[width=\textwidth]{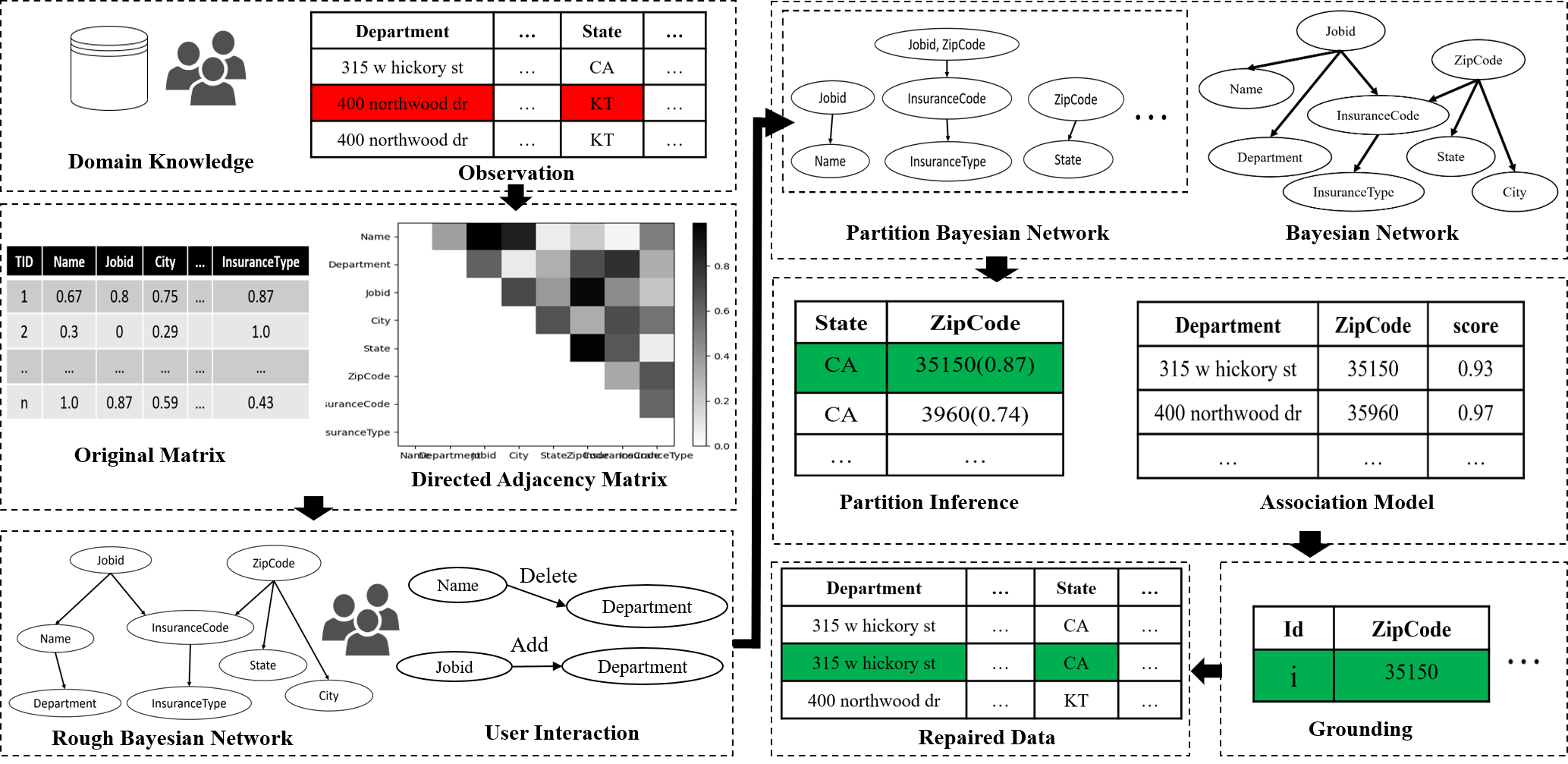}
    \caption{An overview of the \bclean framework.}
    \label{fig:Overview}
\end{figure*} 

A BN is a type of Bayesian method and is presently one of the most effective 
theoretical models in the field of uncertain knowledge representation and
inference~\cite{DBLP:journals/ker/Parsons11a}. A BN is a directed acyclic graph
(DAG) $(N, E, \theta)$, composed of a set of nodes $N$ representing random 
variables, a set of directed edges $E$ indicating the conditional dependencies 
between random variables, and a set of conditional probability tables (CPTs) 
$\theta$ that weight the edges to express the strength of conditional dependency. 
For each random variable, its probability can be computed using the probabilities 
of its node's parents. In case of no parent node, a prior probability is 
expressed by the prior information. Given an observed dataset $D$, the structure 
of a BN and its CPTs can be learned, with nodes representing attributes and edges 
representing dependency relationships. Figure~\ref{fig:Overview} illustrates the 
structure of a BN learned from the dataset in Table~\ref{tab:ExampleDataset}. 
The probability of a tuple $T = (t_1, \ldots, t_m)$ is 
\begin{align*}
    \Pr{t_1, \ldots, t_m} = \prod \limits_{A_i \in N} \Pr{T[A_i] = t_i | T[A_{j}] = t_{j}, j \in Ans(i)}, 
\end{align*}
where $Ans(i)$ represents the set of ancestor nodes influencing $A_i$. 
For nodes without any ancestor nodes, 
$\Pr{T[A_i] = t_i | T[A_{j}] = t_{j}, j \in Ans(i)}$ is determined by the 
prior probability of $A_i$, which can be inferred from $D$.


\section{Modeling in BClean}
\label{sec:framework}
An overview of \bclean is depicted in Figure~\ref{fig:Overview}. \bclean accepts
as input a dataset $D$ as observation, along with UCs provided by users. 
\bclean executes data cleaning in two steps: 

\noindent\textbf{Probabilistic modeling.} 
We employ graphical lasso~\cite{DBLP:conf/nips/WuSD19} to generate a covariance 
matrix and a directed graph structure learning method~\cite{raskutti2018learning} 
to generate a BN. Inference on an erroneous dataset $D$ can exacerbate deviation in 
the results. To counter this issue, we leverage a compensatory score. This 
score considers value frequency (i.e., for each attribute $A_i$, we collect the 
observations for each value in $dom(A_{i})$ as its frequency) and pairwise 
attribute correlation.

\noindent\textbf{Inference grounding and pruning.}
We generate candidate values for data cleaning by iterating through the domain 
values of each attribute and compute the probabilities of the candidate values. 
Considering the number of candidate values can be vast, we curtail unnecessary 
candidate generation through three techniques. First, we partition the BN by 
capitalizing on the Markov property~\cite{DBLP:journals/ker/Parsons11a}. Second, 
based on the partitioned BN, we prune the domain of each attribute and discard 
values that obviously do not conform to the domain semantics. Third, we use the 
compensatory score to perform preliminary detection and inference for the 
variables likely to be correct.


\begin{figure}
    \centering
    \includegraphics[width=0.6\linewidth]{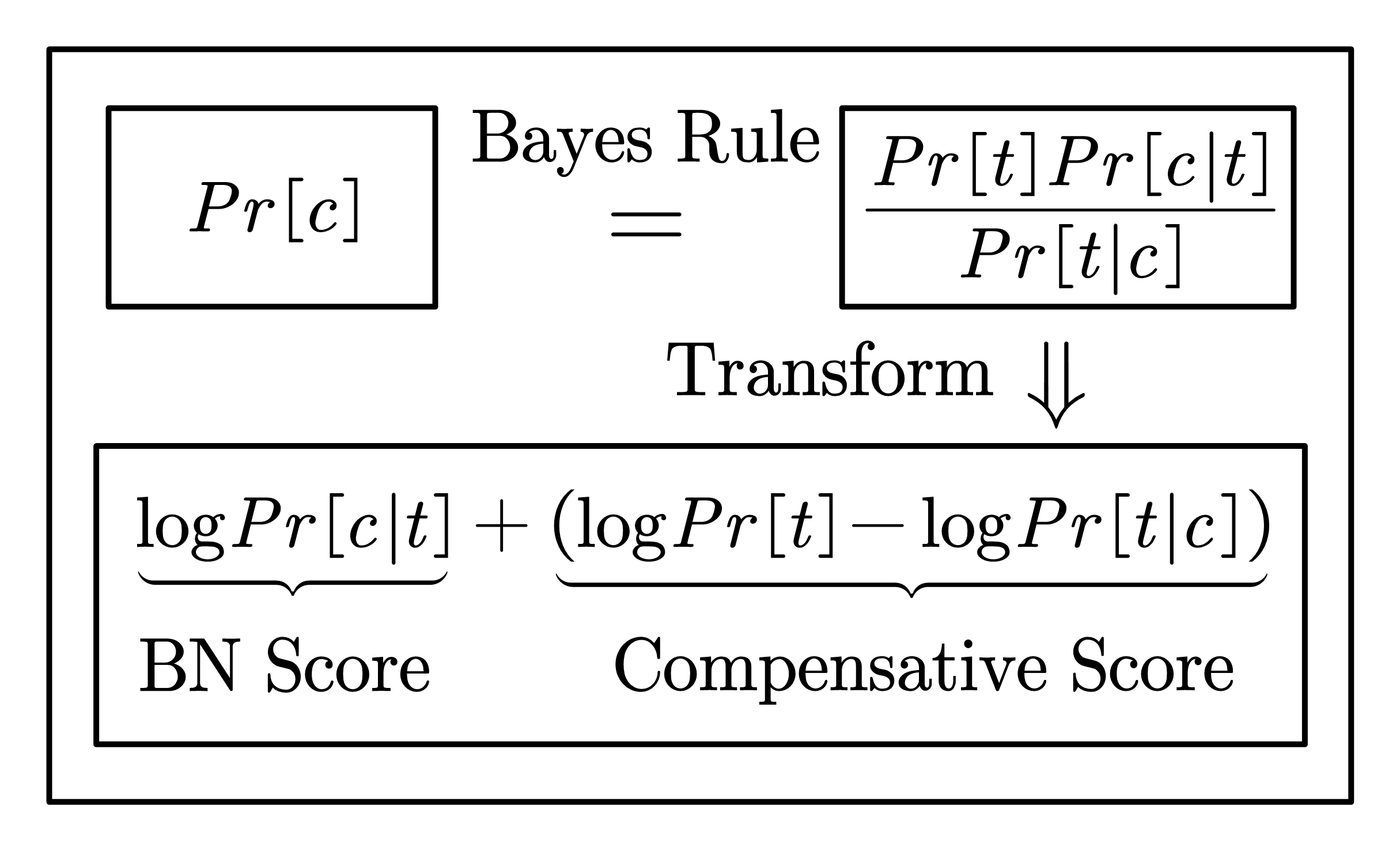}
    \caption{\bclean's data-driven model.}
    \label{fig:Flow_Chart}
\end{figure} 

We represent the data cleaning task as an MAP problem to determine the optimal 
data distribution. Figure~\ref{fig:Flow_Chart} illustrates the modeling 
procedure of \bclean. Given $D$, our objective is to solve an MAP problem to 
derive a cleaned dataset $D^*$. For each $T_{i}$ in $D$ and each attribute $A_j$, 
we infer the most probable candidate value $c^*$ in $dom(A_{j})$. To calculate 
the probability of each candidate value, we leverage the other attributes of 
$T_{i}$ and apply the following transformation via Bayes' rule: 
\begin{align*}
    c^{*} = \arg \max_{c \in dom(A_j)} \frac{\Pr{t} \Pr{c|t}}{\Pr{t|c}}, \text{ subject to } UC(c) = 1,
\end{align*}
where $t$ denotes the observed values of attributes other than $A_j$ in $T[i]$, 
i.e., $t = T_{i}[A_1, \ldots, A_{j - 1}, A_{j + 1}, \ldots, A_m]$. By transforming 
it to logarithmic form, we obtain
\begin{align}
  \begin{split}
    c^{*} = & \arg \max_{c \in dom(A_j)} (\log{\Pr{c|t}} + \log{\Pr{t}} - \log{\Pr{t|c}}), \\
            & \text{ subject to } UC(c) = 1.
  \end{split}
  \label{eq:TransformMultiToAdd}
\end{align}
The terms in the above equation can be divided into two parts: $\log{\Pr{c|t}}$ and 
$\log{\Pr{t}} - \log{\Pr{t|c}}$. The first term can be obtained from the BN using 
the nodes having an edge to $A_j$. While the second term cannot be directly computed 
from the BN, we compute it with a compensatory scoring model, whose details will be 
presented in the next section.

Algorithm~\ref{alg:AlgorithmOne} outlines the procedure of \bclean. It iterates 
through all rows and columns in the observed dataset $D$ and infers the most probable 
value for each cell. For each cell, $c^*$ is initialized as the original value 
$T_i[A_{j}]$. Then, for each candidate value that satisfies UCs, we compute its 
probability using the BN and the compensatory scoring model 
(Equation~\ref{eq:TransformMultiToAdd}), and update $c^*$ if the probability is higher. 
All the $c^*$ values constitute the cleaned dataset $D^*$, which is returned by the 
algorithm.

\begin{algorithm}[t]
  \small
  \caption{\bclean}
  \label{alg:AlgorithmOne}
  \SetFuncSty{textsf}
  \SetArgSty{textsf}
  \Input{Observed dataset $D$ with $n$ rows and $m$ columns, Bayesian network $BN$.}
  \Output{Cleaned dataset $D^{*}$.}
  \For{$i = 1$ to $n$}{
    \For{$j = 1$ to $m$}{
        \State{$c^{*} \gets T_i[A_{j}]$}
        \State{$p^{*} \gets \log(BN[A_j](c^{*})) + \log(CS[A_j](c^{*}))$}
        \For{$c \in dom(A_{j})$ such that $UC(c) = 1$}{
            \State{$p \gets \log(BN[A_j](c)) + \log(CS[A_j](c))$}
            \If{$p > p^{*}$}{
                \State{$c^{*} \gets c$, $p^{*} \gets p$}
            }
        }
        \StateCmt{$D_{i}^{*}[A_j] \gets c^{*}$}{clean ($i$-th row, $j$-th column)}
    }
  }
  \Return{$D^{*}$}
\end{algorithm}

\section{Bayesian Network Construction}
\label{sec:network}
We introduce our technique for constructing a BN. The 
BN partitioning technique will be described in Section~\ref{sec:optimization}.

The problem of constructing a BN from a given dataset is 
NP-hard~\cite{chickering1995learning}. Therefore, an exhaustive search 
algorithm that traverses every structure is not practical for large 
datasets. Many existing approaches resort to heuristics to find a 
feasible solution. For instance, hill-climbing-based approaches, such as MMHC~\cite{DBLP:journals/ml/TsamardinosBA06} provided in the 
Pgmpy toolkit~\cite{ankan2015pgmpy}, add one edge at a time and evaluate 
its score to determine each edge and its direction. However, such 
approaches often converge to a local optimum, limiting their 
effectiveness. Other typical approaches include tree search, which 
necessitates specifying the root state, and the PC 
algorithm~\cite{Spirtes1991AnAF}, which requires a conditional 
independence hypothesis given by the user.

In \bclean, we construct a BN from the dataset $D$, as depicted in 
Figure~\ref{fig:Overview}. Automatic BN construction is challenging because 
it either requires accurate data or prior knowledge of the network (e.g., 
as specified using PPL~\cite{DBLP:conf/aistats/LewASM21}). However, $D$ is 
often unclean and sparse. Our objective is to create a supportive system 
that obviates the need for users to initiate BN construction from scratch. 
This makes the aforementioned heuristic approaches ill-suited to our use 
case. To mitigate this, we leverage a structure learning algorithm that 
supports domain knowledge to construct an approximate BN. Our strategy 
involves extending the FDX method~\cite{DBLP:conf/sigmod/ZhangGR20} with 
error tolerance to generate an \emph{inverse covariance matrix} using 
\emph{graphical lasso}~\cite{DBLP:conf/nips/WuSD19}, and then employing the 
inverse covariance matrix to create a BN skeleton. Given that the BN 
skeleton could be noisy and incomplete, we offer a user-interaction feature 
enabling users to view and manually adjust the BN. It is worth noting that 
there are alternative methods to construct a BN from $D$. However, we have 
chosen the solution outlined above due to its ability to tolerate errors in 
$D$ and its superior empirical performance. Next, we describe the details. 


Our method, based on in statistical modeling, operates directly on $D$ and 
models the distribution by which clean data is generated. To model the 
relationships between attributes, we utilize the FDs within the dataset. An 
FD $X \rightarrow Y$ states that $T_{i}[Y] = T_{j}[Y]$, if 
$T_{i}[X] = T_{j}[X]$. Given the presence of errors in the dataset, the 
application of strict FDs lacks flexibility. Therefore, we propose to soften 
the FDs by introducing a similarity measure: for any pair of tuples 
$T_{i}, T_{j} \in D$, we denote $Sim(T_{i}[X], T_{j}[X])$ as the similarity 
between two values. The similarity ranges between 0 and 1, and serves as 
probability in our modeling. For numerical data, we use 
$\frac{|T_i[X] - T_j[X]|}{(|T_i[x]| + |T_j[x]|)/2}$, and for string data, we 
use unit-cost edit distance normalized by string lengths as follows.
\begin{align*}
  Sim(T_{i}[X], T_{j}[X]) = 1 - \frac{2 \cdot ED(T_{i}[X],T_{j}[X])}{len(T_{i}[X]) + len(T_{j}[X])}, 
\end{align*}
where $ED$ denotes unit-cost edit distance (i.e., Levenshtein distance) and 
$len$ denotes string length. For example, as illustrated in the \kw{Department}
attributes of Tuples 1 and 3 in Figure~\ref{tab:ExampleDataset}, the features 
obtained here are represented by 1 because they refer to the same entity. If 
an FD is applied, it would be 0, whereas our softened method reports a 
similarity of 0.86, thereby tolerating errors in the dataset.

We compute pairwise similarity for the attributes in each tuple, considering 
these similarity values as observations from a multivariate Gaussian 
distribution. The graphical lasso~\cite{DBLP:conf/nips/WuSD19} is subsequently 
utilized to calculate the covariance matrix $\Sigma$ of the underlying 
distribution. We decompose the inverse covariance matrix 
$\Theta = \Sigma^{-1}$, adhering to the method typically used in learning the 
structure of linear models~\cite{loh2014high,raskutti2018learning} and employed 
in FDX~\cite{DBLP:conf/sigmod/ZhangGR20}:
\begin{align*}
    \Theta = \Sigma^{-1} = (I - B) \Omega (I - B)^{T}, 
\end{align*}
where $\Theta$ denotes the inverse covariance matrix for pairwise attributes in 
$\mathcal{A}$, $I$ is the identity matrix, and $B$ is the autoregression matrix 
of the model, which is exactly the adjacency matrix that stores the weights of 
the edges of the BN skeleton. Moreover, we set a weight threshold, thereby 
retaining only edges with weights exceeding the threshold. 

For example, for dataset $D$ in Table~\ref{tab:ExampleDataset}, we compute 
pairwise similarities, as depicted in the second part of Figure~\ref{fig:Overview} 
labeled \kw{original matrix}. Following this, we compute the inverse covariance 
matrix and decompose it to obtain the adjacency matrix of the BN skeleton. 
$(ZipCode, InsuranceCode)$ becomes a directed edge since its weight in the 
adjacency matrix surpasses the threshold.

Upon constructing the BN skeleton, we conduct parameter learning to estimate the 
joint probability of attributes and derive the CPTs concerning the observations. 
It is important to note that during the construction of the BN skeleton, we are 
oblivious to any errors in the observations, and our BN construction models errors 
as part of the distribution.

Given that the BN skeleton may be noisy, we offer an interface through which users 
can view and manually modify the automatically constructed skeleton to generate 
the BN employed for inference. Users can add or remove edges in the BN, as depicted 
in Figures~\ref{fig:Overview} (f) -- (g), or merge nodes, as shown in 
Figures~\ref{fig:Overview} (g) -- (h). For nodes to be merged, if all of them have 
an edge from/to node $A_j$, the edges from/to $A_j$ will be merged into one, as 
displayed in the incoming edge to \kw{InsuranceCode} in Figure~\ref{fig:Overview} (h). 
Other edges from/to the nodes to be merged will be removed after merging. Since edges 
are associated with CPTs, the CPTs need to be updated if users modify the BN. For 
efficiency, we only recalculate the CPTs for the attributes involved in the 
modification instead of all attributes.

\textbf{Remarks.} 
To use the FDX method, we first sort tuples according to each attribute, and only compute similarities and check equality within two adjacent tuples. As such, we do not need to compute each tuple pair in $D$. We construct the Bayesian network by using the FDX method and extend it to support fuzzy matching, e.g., edit similarity. The time complexity is $(nm \log n + nm + r\size{B})$, where $r$ is the number of iterations of graphical lasso learning, and $\size{B}$ is the size of the autoregression matrix.


\section{Bayesian Inference with Compensatory Score}
\label{sec:association}
Upon construction of the BN, we design an inference method incorporating 
a compensatory strategy to identify the most probable candidate value for 
each attribute of a tuple $T$. Traditional approaches to BN inference 
involve the iterative cleaning of data via MAP estimation. However, such 
a method is often ineffective as most real-world datasets are unclean and 
the BNs constructed from them lack accuracy. If a repair is incorrect, the 
error propagates to the next step, rendering subsequent repairs based on 
it likely invalid. We refer to this problem as error amplification. Recall 
that we partition the logarithmic form of $\log \Pr{c}$ into two terms, 
$\log \Pr{c|t}$ and $\log \Pr{t} - \log \Pr{t|c}$, where $\log \Pr{c|t}$ 
can be computed through BN inference via CPTs. To minimize the errors 
produced by $\log \Pr{c|t}$, we consider the second term as a compensatory 
score $\kw{Score}\kw{comp}$, i.e., 
$\kw{Score}\kw{comp} = \log{\Pr{t}} - \log{\Pr{t|c}}$.

\begin{example}\label{eg:comp-motiv}
  Consider the \kw{BN} of the relational data in Table~\ref{tab:ExampleDataset}, 
  where we aim to populate the missing value for the attribute \kw{Department} of 
  $t_6$. Owing to the unclean data, we derive 
  $\Pr{\text{``400 nprthwood dr''} | t_6} = 0.54$ and 
  $\Pr{\text{``400 northwood dr''} | t_6} = 0.51$. Without $\kw{Score}_\kw{comp}$, 
  ``400 nprthwood dr'' would be erroneously filled into the cell for 
  \kw{Department} of $t_6$.
\end{example}

Computing $\kw{Score}\kw{comp}$ directly is difficult, as it necessitates 
the knowledge of \emph{$\Pr{c^*}$}, thereby inverting the dependency between $c^*$ 
and $t$. Given that $\Pr{t}$ is a constant value once the \kw{BN} is 
constructed and fixed, a lower value of $\Pr{t|c}$ suggests $\Pr{c}$ is closer to 
$\Pr{c^*} = 1$. We approximate the computation of the compensatory score as the 
correlation between $c$ and $t$:
\begin{align*}
    \kw{Score}_\kw{comp}(c, t) \approx \kw{Score}_{\kw{corr}}(c, t).
\end{align*}

To explain this approximation, 
intuitively, if a candidate value $c \in \kw{dom}(A_i)$ of a tuple $T$ is $c^*$, $c$ 
is likely to exhibit a strong dependency to other attributes, i.e., coexist with 
other values frequently. We assume that erroneous values in real-world datasets are 
rare. For example, consider an error $c'$ which only appears in one tuple $T$. In this 
case, $\Pr{t|c} = 1$ and $-\log \Pr{t|c}$ attains the minimum value, i.e., 0. Thus, 
modeling the correlation between $c$ and $t$, i.e., $\kw{Score}\kw{corr}$, aligns 
with the same objective as $\kw{Score}\kw{comp}$.

Inspired by Bayeswipe~\cite{DBLP:journals/jdiq/DeHMCK16}, we adopt a 
straightforward yet effective idea to compute $\kw{Score}\kw{corr}$ by counting the 
occurrences of $(c, t)$. Accordingly, 
$\kw{Score}\kw{corr} = \frac{\kw{count}(c, t)}{|D|}$. Since this scoring suffers 
from sparsity and tends to be inaccurate for infrequent observations, to mitigate 
this, we separately consider the correlation between each attribute value of $t$ and 
$c$ and accumulate them. If more values in $t$ have a high correlation with $c$, 
then $\kw{Score}\kw{corr}(c, t)$ is likely to be significant. Since $c = T_i[A_j]$ 
and $t = T_i[A_1, \dots, A{j-1}, A_{j+1}, \dots, A_m]$, we define 
$\kw{Score}_\kw{corr}$ as follows:
\begin{align}\label{eq:score}
      \kw{Score}_\kw{corr}(c, t, A_j) = \sum\limits_{A_k \in \mathcal{A} \setminus \set{A_j} \land e = t[A_k]} 
      \kw{corr}(c, e, A_j, A_k),
\end{align}
where $\kw{corr}(c, e, A_j, A_k)$ denotes the correlation between value $c$ of 
attribute $A_j$ and value $e$ of attribute $A_k$. Due to the presence of noise, 
instead of directly counting the occurrence of $(c, e)$ as their correlation such 
that $\kw{corr}(c, e, A_j, A_k) = \frac{\kw{count}(c, e, A_j, A_k)}{|D|}$, we 
incorporate the UCs provided by users and introducing the concept of confidence 
(\kw{conf}) for each tuple $T_i$. We assign different weights (i.e., confidences) to 
tuples. A tuple $T_i$ with a higher confidence value suggests that it is more 
accurate, and the correlations among its values are more reliable. We define 
$\kw{conf}(T_i)$ as follows, where $UC(\cdot)$ is a function to check each attribute 
value against user constraints and output either 0 or 1:
\begin{align}\label{eq:conf}
    \kw{conf}(T_{i}) = \max \big{ \{ }0, \frac{\sum\limits_{e \in T_i}{\mathbf{1}_{\{UC(e) = 1\}}} - \lambda \cdot \sum\limits_{e \in T_i}{\mathbf{1}_{\{UC(e) = 0\}}} }{|T_i|} \big{ \} },  
\end{align}
where $\lambda$ is a non-negative parameter used to penalize the erroneous values 
that violate the UCs, and $\mathbf{1}$ is the indicator function. Then, we define 
$\kw{corr}(c, e, A_j, A_k)$ as 
\begin{align*}
  \kw{corr}(c, e, A_j, A_k) &=
  \frac{ \sum_{T \in \Omega}{ (\mathbf{1}_{\kw{conf}(T) \geq \tau}  - \beta \mathbf{1}_{ \kw{conf}(T) < \tau }})}{|D|}, \\
  \Omega &= \{ T' | T' \in D \land (c,e) = (T'[A_j],T'[A_k]) \}. 
\end{align*}
In the above equation, if the confidence of a tuple $T \in D$ is no less than a 
predefined constant $\tau$, $T$ is considered reliable and the combinations of its 
attribute values (i.e., $(T[A_j], T[A_k])$) contribute to the computation of 
$\kw{corr}$. Otherwise, $T$ is considered unreliable, and we impose a penalty $\beta$ 
such that $\kw{corr}$ would decrease once $T$ is taken into account.

Since we incorporate UCs in Equation~\ref{eq:conf}, more violations of UCs will 
reduce $\kw{Score}\kw{corr}$. It is important to note that the inference process is 
a competition among the candidate values in the domain, comparing their scores. In other 
words, the relative order is significant, not the scores themselves. Therefore, when 
using $\kw{Score}\kw{corr}$, it is not necessary to require $\kw{Score}\kw{corr}$ to be 
equal to $\kw{Score}\kw{comp}$. 

To efficiently compute $\kw{Score}\kw{corr}$ for all attributes, we design a 
compensatory score computation algorithm  
(Algorithm~\ref{alg:AlgorithmTwo}). We scan each attribute value in $D$ and compute the 
correlation with other attribute values (lines 2 -- 12). Specifically, we check each 
tuple $T \in D$ (line 2) and compute its confidence $\kw{conf}(T)$ using 
Equation~\ref{eq:conf}. Then, we enumerate all pairs of attributes $(A_i, A_j)$ with two 
cases: if $\kw{conf}(T)$ is no less than the predefined threshold $\tau$, we update 
$\kw{corr}$ by accumulating the counts of $(A_i, A_j)$; otherwise, we impose a penalty 
$\beta$ to decrease $\kw{corr}$. Finally, we compute and store the correlation values of 
all pairs of values in $D$, and $\kw{Score}\kw{corr}$ could be readily computed using 
Equation~\ref{eq:score}.

\begin{algorithm}[t]
  \small
  \caption{compensatory score computation} 
  \label{alg:AlgorithmTwo}
  \LinesNumbered
  \KwIn{Observed dataset $D$, user constraints $UC(\cdot)$, confidence threshold $\tau$, penalty parameter $\beta$;}
  \KwOut{compensatory scores $\kw{corr}$.}
  $\kw{corr} = \{\}$\; 
  \For{$T \in D$}{
    $p :=\emptyset$\;
    Compute $\kw{conf}(T)$ using Equation~\ref{eq:conf}\;
    \For{$A_{i} \in A$}{
        $c := T[A_{i}]$\;
        \For{ $A_{j} \in A / A_{i}$}{
            $v :=T_{i}[A_{k}]$\;
            \eIf{$ \kw{conf} \geq \tau$}{
                $\kw{corr}[\langle{c, v, A_i, A_j}\rangle ] = \kw{corr}[\langle{c, v, A_i, A_j}\rangle ] + 1$\;
            }{
                $\kw{corr}[\langle{c, v, A_i, A_j}\rangle ] = \kw{corr}[\langle{c, v, A_i, A_j}\rangle ] - \beta$\;
            }
        }
    }
  }
  \Return{$\kw{corr}$}
\end{algorithm}

\begin{example}
  Continuing with Example~\ref{eg:comp-motiv}, we set $\lambda = 0.25$, 
  $\beta = 2$, and $\tau = 0.75$. Consider a $UC(\cdot)$ as a spell checker
  that returns 1 if the spelling is correct and 0 otherwise. 
  $c = \texttt{400 nprthwood dr}$ in $t_5$ violates the $UC$, and its 
  $\kw{conf}(t_5) = 0 < \tau$ and 
  $\kw{Score}_\kw{corr}(c, t_5, \kw{Department}) = -0.31$. 
  $c = \texttt{400 northwood dr}$ in $t_6$ does not violate the UC, and 
  its $\kw{conf}(t_6) = 0.79 \geq \tau$. $c = \texttt{400 northwood dr}$ 
  in $t_4$ does not violate the UC, and its 
  $\kw{Score}_\kw{corr}(c', t_4, \kw{Department}) = 0.13$. 
  Eventually, we have $\Pr{\texttt{400 northwood dr} | t_6} = 0.64$ and 
  $\Pr{\texttt{400 nprthwood dr} | t_6} = 0.23$. Upon executing Bayesian 
  inference with a compensatory score, the \kw{Department} of $t_6$ is 
  filled with \texttt{400 northwood dr}.
\end{example}

\textbf{Remarks.} 
While our idea is inspired by Bayeswipe~\cite{DBLP:journals/jdiq/DeHMCK16}, its 
realization is completely different. Instead of focusing on total loss as Bayeswipe 
does, our correlation-based score $\kw{Score}_\kw{corr}$ is derived from the 
co-occurrence of attribute-value pairs in $D$ and the confidences of tuples. In 
real-world data, clean data constitute the majority and exhibit dependency and 
correlation. If a tuple is clean, its correlation score is high as it is likely to 
share common attribute values with a few other tuples. Conversely, if a tuple 
contains errors, it is less likely to share common values with other tuples, 
resulting in a very low correlation score for $t$ observed. Thus, we approach this 
as a missing value prediction task in the correlation-based score. For a tuple 
$T_{i} \in T$, the value of every cell ${T_{i}[A_{i}] | 0 \le i \le m}$ is more 
likely to be clean for higher co-occurrence times with the remaining observations 
${T_{i}[A_{j}] | 0 \le j \le m \land j \neq i}$ of the column. We first create a 
co-occurrence dictionary where all candidate values $c$ are stored along with the 
count of their co-occurrence with every possible value from all other columns. For a 
specific tuple $t$, $\kw{Score}{\kw{corr}}(c, t)$ will collect the count of cell 
observations from different columns shown in $T{i}$ as features. The norm of the 
feature is computed as a weighted score, where the weights are the counts of 
co-occurrence of $(c, t)$. The distance between an observation and a candidate value 
is matched with the weighted score. Candidate values with higher scores have a 
greater likelihood of being the final ground truth. The time complexity of 
Bayesian inference is dominated by compensatory score computation, which is $O(nm^2)$.


\section{Bayesian Inference Optimization}
\label{sec:optimization}
When the dataset is large, the number of variables in BN increases, rendering 
the BN inference cost prohibitive. To address this issue, a partition inference 
mechanism that conducts inference on nodes in BN, only interacting with those 
nodes within one hop in the Markov blanket. In addition, propose two pruning 
strategies: a tuple pruning strategy that identifies the cells in the dataset 
requiring repair and bypasses those that are largely clean, and a domain pruning 
strategy that eliminates values in domains that are clearly not candidate values 
for repair. 


\subsection{Partitioned Bayesian Inference}
Bayesian inference methods can be either exact inference such as variable 
elimination and belief propagation, or approximate inference such as Gibbs 
sampling. Unlike existing techniques primarily concentrate on the global 
distribution of BN, according to the Markov property, we can treat the 
inferred node $A_{j}$ as a state in BN, while disregarding the set of nodes 
that are not directly connected to $A_{j}$. This process essentially 
transforms a BN into several sub-networks. Each sub-network computes the 
partitioned distribution pertaining to the dividing node $A_{j}$, with all 
other nodes considered as observations. During Bayesian inference on each 
node, only nodes and edges within its sub-network participate in the 
calculation.

\noindent\textbf{BN partitioning.}
The variable elimination strategy for exact inference necessitates the inclusion of 
all precursor states of $A_{j}$ in the topology, according to the inference rules 
of BN. However, the states of precursor nodes might have been altered in previous 
repairing steps. As such, the prior of $A_{j}$ may not be the original one, and 
inferring from the first node could lead to higher costs when inferring every $A_{j}$. 
If the modified states are erroneous, these errors could propagate to the inference of 
$A_{j}$. To avoid this erroneous propagation and to expedite the inference, we 
partition BN into $l$ sub-networks (we abuse the notation of a set of attributes to 
denote a sub-network): 
$\set{\mathcal{A}_\kw{joint}^{(1)}, \mathcal{A}_\kw{joint}^{(2)}, \ldots, \mathcal{A}_\kw{joint}^{(l)}}$, 
following the Markov blanket. $\mathcal{A}^{(i)}\kw{joint}$ ($1 \leq i \leq l$) denotes the 
$i$-th sub-network containing the inferred node $A{j}$, alongside its one-hop parent 
nodes $\mathcal{A}^{(i)}\kw{parent}$ and child nodes $\mathcal{A}^{(i)}\kw{child}$. Formally,
\begin{align*}
    \mathcal{A}^{(i)}_\kw{joint}=\mathcal{A}^{(i)}_\kw{parent} \cup \set{A_{j}} \cup \mathcal{A}^{(i)}_\kw{child}.
\end{align*}
Multiple sub-networks might intersect at a node $A{k}$, but $A_{k} \in \mathcal{A}^{(i)}_\kw{joint}$ 
does not affect other sub-networks. 

\noindent\textbf{Bayesian inference.}
In the partitioned BN, nodes can be categorized into two types. One is isolated nodes, 
$\mathcal{A}_\kw{iso}$, which do not connect to other nodes. The other is joint nodes, 
$\mathcal{A}_\kw{joint} \in \kw{BN}\kw{sets}$, which connect to other nodes. For each 
isolated node, the CPT is modeled as uniform distribution, under the assumption that 
candidate values are uniformly distributed in the domain. For each joint node, all 
observations are known; thus, the probability distribution of candidate values is 
directly obtained from the CPT. The set of connected nodes, $\mathcal{A}_\kw{connected}$, 
includes parent nodes $\mathcal{A}_\kw{parent}$ and child nodes 
$\mathcal{A}_\kw{child}$, with CPTs given by $\Pr{\mathcal{A}_{j}|A_\kw{parent}}$ and 
${\Pr{\mathcal{A}_\kw{child}|A_{i}}}$. In accordance with the properties of BN, we 
treat $A_{j}$ as known, denoting every value in the domain as a candidate. As such, 
$\mathcal{A}_\kw{parent}$ and $\mathcal{A}_\kw{child}$ are independent. Formally,
\begin{align*}
    \Pr{A_{j}|\mathcal{A}_\kw{connected}} = \Pr{A_{j}|\mathcal{A}_\kw{parent}}\Pr{\mathcal{A}_\kw{child}|A_{j}}.
\end{align*}

\subsection{Pruning Strategies} 
Data cleaning on $D$ with values drawn from the domain of their corresponding 
attribute, and the inference cost is capped by the number of cells and the 
number of candidate values. Consequently, we introduce tuple and domain pruning 
strategies to respectively prune cells that do not need inspection and candidate 
values that are not correct answers.

\noindent\textbf{Tuple pruning.}
Given a tuple $T$ and an attribute $A_i$, we design a filtering mechanism to 
determine whether $T[A_i]$ should be inferred. The underlying intuition is that 
if $T[A_i]$ co-occurs with other values of $T$ more frequently, indicating a 
stronger correlation, then $T[A_i]$ is more likely to be correct and $T[A_i]$ has 
a lower priority to be inferred. We define a function $\kw{Filter}(T, A_{i})$ and 
a threshold $\tau_{\kw{clean}}$ to tell whether $T[A_i]$ needs inference. Formally,
\begin{align*}
  \kw{Filter}(T, A_{i}) = \frac{1}{m - 1} \sum_{A_{j} \in \mathcal{A} \setminus \set{A_{i}}} \frac{\kw{count}(T[A_{i}], T[A_{j}])}{\kw{count}(T[A_{j}])}.
\end{align*}

In the data cleaning process, we first compute $\kw{Filter}(T, A_i)$. If the result 
is not less than $\tau_\kw{clean}$, we consider $T[A_i]$ as relatively reliable and 
omit the BN inference in the current iteration; otherwise, $T[A_i]$ is deemed 
obscure and requires repair. It is important to note that tuple pruning differs 
from standard error detection. Even though both can identify erroneous values, tuple 
pruning prioritizes cells with the fewest conflicts.


\noindent\textbf{Domain pruning.}
We treat each sub-network as an independent semantic space and perform domain 
pruning from the semantic perspective, akin to the cloze test task in natural 
language processing.

In this semantic space, we take the dividing variable as the fill-in-the-blank 
variable, with all variables excluding the dividing variable acting as the 
context, and compute each answer under these semantics. For every value 
$v \in \kw{dom}(A_{i})$, we assign a weight using TF-IDF. Formally,
\begin{align*}
    \kw{sorce}(v)
    &= \kw{TF}(v, \kw{context})\kw{IDF}(v, D)\\
    &= \kw{context}(v)\ \cdot \log(\frac{|D|}{1+\kw{count}(v, D)}),
\end{align*}
where $\kw{context}(v)$ denotes the number of sub-networks filled with $v$. These 
semantics treat different sub-networks as distinct contexts. The more frequently 
${v}$ appears in a specific semantic and the less frequently ${v}$ appears in other
semantics, the more likely ${v}$ is to become the ground truth in that semantic. 
For every non-semantic $v \in \kw{dom}(A_{i})$, its score is significantly lower, 
possibly even zero. Each sub-network only computes a few candidates likely to be 
the ground truth during inference.

\section{Experiments}
\label{sec:exp}

\subsection{Experiment Setup}
\label{sec:exp-setup}

\noindent{}\textbf{Datasets.}
We use six benchmark datasets of varying sizes and error types as summarized in 
Table~\ref{DatasetSummary}. 
\begin{itemize} [leftmargin=*]
  \item \textbf{\hospital}: This is a dataset derived from \cite{DBLP:journals/pvldb/RekatsinasCIR17,DBLP:conf/icde/ChuIP13}, 
  primarily comprising details of various hospitals such as \kw{HospitalName} 
  and \kw{Address}. Approximately 5\% of the data contains errors, with ground 
  truth available for all cells. This dataset exhibits significant duplication 
  across cells and possesses substantial causality between attributes.
  \item \textbf{\flights}: Used in \cite{DBLP:journals/pvldb/RekatsinasCIR17} 
  and collected from \cite{li2015truth}, this 
  dataset contains departure and landing times of different airlines recorded on 
  various websites. Ground truth information is available for part of the cells. 
  \item \textbf{\soccer}: This dataset contains profiles of football players, 
  including attributes like \kw{Name}, \kw{Date of Birth}, \kw{City}, etc. The 
  clean version of \soccer is collected from \cite{rammelaere2018explaining}. 
  Errors account for approximately 5\% of all data.
  \item \textbf{\beers}: This dataset used in  
  \cite{mahdavi2019raha, DBLP:journals/pvldb/MahdaviA20} contains two numerical 
  attributes, \kw{ounces} and \kw{abv}. We acquire both the clean and dirty 
  versions from the same source.
  \item \textbf{\inpatient}: This dataset of inpatient profiles was collected from 
  the CMS~\cite{cms}. 
  \item \textbf{\facilities}: This dataset contains information on medical 
  enterprises collected from the CMS~\cite{cms}. 
\end{itemize}

\noindent{}\textbf{Error Injection.}
In line with the error injection approach utilized in the benchmarks (e.g., \hospital) 
by \rahabaran~\cite{mahdavi2019raha,DBLP:journals/pvldb/MahdaviA20} and 
\holoclean~\cite{DBLP:journals/pvldb/RekatsinasCIR17}, by default, 
we categorize errors into three types: typos (T), missing values (M), inconsistencies 
(I). For T, we modify the original value by randomly adding, deleting, or replacing a 
character. For M, we randomly replace a value with NULL. For I, we interchange two 
values from the domains of two columns or a specific column. Their frequencies do not 
exhibit a significant difference, e.g., in the \inpatient dataset, we observe 1210, 
1800, and 1480 instances of T, M, and I errors, respectively. 
In addition to these default error settings, 
we also evaluate swapping value errors (S), which are generated by swapping values 
within the same attribute, i.e., the same domain.

\begin{table*}
  \small
  \caption{Statistics of datasets. Error types include typos (T),
  missing values (M), inconsistency (I), and swapping value errors (S).} 
  \vspace{-2ex}
  \centering
  \resizebox{\linewidth}{!}{%
  \begin{tabular}{c c c c c c c c}  
    \hline
    Dataset  & Size (\#rows, \#columns, \#cells) & Noise rate & Error types & \#UCs (\bclean) & \#DCs (\holoclean) & \#lines of PPL (\pclean) & \# labels of tuples (\raha + \baran) \\
    \hline
    \hospital & (1000, 15, 15k) & $\sim$5\% &T, M, I & 15 & 13 & 51 & 20+20\\

    \flights & (2376, 6, 14k) & $\sim$30\% &T, M & 6 & 4 & 36 & 20+20\\

    \soccer & (200000, 10, 2M) & $\sim$1\% & T, M, I & 10 & 4 & 37 & 20+20\\

    \beers  & (2410, 11, 27k) & $\sim$13\% & T, M, I & 11 & 6 & 26 & 20+20\\

    \inpatient  & (4017, 11, 44k) & $\sim$10\% & T, M, I, S & 11 & 3 & 54 & 20+20\\

    \facilities  & (7992, 11, 88k) & $\sim$5\% & T, M, I, S & 11 & 8 & 48 & 20+20\\ \hline
    
  \end{tabular}
  }
  \label{DatasetSummary}
\end{table*}

        
        
        
        
        
        
        
        
        
        
        
        

\begin{table*}
  \small
  \caption{List of user constraints (UCs).}
  \centering
  \vspace{-2ex}
  \resizebox{\linewidth}{!}{%
  \begin{tabular}{c c}  
    \hline
    DataSet  & UCs (max. and min. length constraints for all textual attributes, not-null value constraints for all attributes.)\\
    \hline
    \hospital & $\wedge$\texttt{([1-9][0-9]\{4, 4\})}[\kw{ProvideNumber}, \kw{ZipCode}]; $\wedge$\texttt{([1-9][0-9]\{9, 9\})}[\kw{PhoneNumber}]; 
    \\
    \flights & \texttt{([1-9]:[0-5][0-9][s][ap].[m].|1[0-2]:[0-5][0-9][s][ap].[m].|0[1-9]:[0-5][0-9][s][ap].[m].)}[sched\_dep\_time, act\_dep\_time, sched\_arr\_time, act\_arr\_time] \\
    \soccer & \texttt{([1][9][6-9][0-9])}[\kw{birthyear}]; \texttt{([2][0][0-9][0-9])}[\kw{season}]\\
    \beers & $\setminus\texttt{d}+\setminus.\setminus\texttt{d}+|(\setminus\texttt{d}+)$[\kw{ounces}, \kw{abv}] \\
    \inpatient & N/A \\
    \facilities &  N/A \\
    \hline
  \end{tabular}
  }
  \label{tab:ucs}
\end{table*}

\begin{table*}[ht!]
  \centering
  \small
  \caption{Precision (P), recall (R), and F1-score (F1) of data cleaning methods.}
  \vspace{-2ex} 
  \resizebox{\linewidth}{!}{%
      \begin{tabular}{|l | c c c | c c c | c c c|  c c c | c c c | c c c|}
      \hline
      \multirow{2}{*}{\textbf{Method}}  & \multicolumn{3}{c|}{\hospital} & \multicolumn{3}{c|}{\flights} & \multicolumn{3}{c|}{\soccer} & \multicolumn{3}{c|}{\beers} & \multicolumn{3}{c|}{\inpatient}
      & \multicolumn{3}{c|}{\facilities}\\
      \cline{2-19}
      \multirow{2}{*}{} & \textbf{P}  & \textbf{R} & \textbf{F1} & \textbf{P}  & \textbf{R} & \textbf{F1} & \textbf{P}  & \textbf{R} & \textbf{F1} &
      \textbf{P}  & \textbf{R} & \textbf{F1} &
      \textbf{P}  & \textbf{R} & \textbf{F1} & \textbf{P}  & \textbf{R} & \textbf{F1} \\
      \hline
      \bcleannouc & \textbf{1.000} & 0.935 & 0.966 & 0.807 & 0.729 & 0.766 & 0.927 & 0.982 & \textbf{0.954} & 0.880 & 0.065 & 0.121 & 0.934 & \textbf{0.883} & \textbf{0.908} & 0.810 & \textbf{0.805} & \textbf{0.807}\\
      \bclean & 0.998 & 0.956 & 0.976 & 0.852 & 0.816 & 0.834 & \textbf{0.928} & 0.979 & 0.952
      & 0.916 & 0.887 & 0.901 & 0.909 & 0.845 & 0.876 & - & - & -\\
      \bcleanII & \textbf{1.000} & \textbf{0.960} & \textbf{0.980} & 0.831 & 0.780 & 0.805 & 0.919 & \textbf{0.986} & 0.951 & 0.948 & \textbf{0.949} & \textbf{0.949} & 0.934 & \textbf{0.883} & \textbf{0.908} & 0.810 & \textbf{0.805} & \textbf{0.807}\\
      \bcleanIII & 0.997 & 0.903 & 0.948 & 0.830 & 0.784 & 0.807 & 0.845 & 0.931 & 0.885 
      & 0.948 & 0.882 & 0.914 & 0.929 & 0.791 & 0.855 & 0.753 & 0.730 & 0.741\\
      \pclean & \textbf{1.000} & 0.927 & 0.962 & 0.907 & \textbf{0.884} & \textbf{0.895} & $0.184^{\dagger}$ & $0.672^{\dagger}$ & $0.289^{\dagger}$ 
      & $0.028^{\dagger}$ & $0.028^{\dagger}$ & $0.028^{\dagger}$ & $0.576^{\dagger}$ & $0.460^{\dagger}$ & $0.512^{\dagger}$ & - & - & -\\
      \holoclean & \textbf{1.000} & 0.456 & 0.626 & 0.742 & 0.352 & 0.477 & - & - & - 
      & \textbf{1.000} & 0.024 & 0.047 & 0.966 & 0.219 & 0.357 & \textbf{1.000} & 0.612 & 0.759 \\
      \rahabaran & 0.971 & 0.585 & 0.730 & 0.829 & 0.650 & 0.729 & 0.768 & 0.103 & 0.182 
      & 0.873 & 0.872 & 0.873 & 0.643 & 0.442 & 0.524 & 0.499 & 0.309 & 0.382  \\
      \garf & \textbf{1.000} & 0.556 & 0.715 & \textbf{0.968} & 0.012 & 0.024 & 0.667 & 0.534 & 0.583 
      & 0.973 & 0.011 & 0.021 & \textbf{0.971} & 0.091 & 0.166 & 0.963 & 0.281 & 0.435  \\
      \hline
      \end{tabular}
  }      
  \begin{tablenotes}
      \footnotesize
      \item $\dagger$ We invite people familiar with \pclean to author the data models in PPL. 
      \item $-$ Out-of-memory, out-of-runtime (24h), or no repairs.
  \end{tablenotes}
  \label{BaselineResult}
\end{table*}

\noindent{}\textbf{Methods.}
We compare the following methods. 
\begin{itemize} [leftmargin=*]
  \item \bclean: This is the basic version of \bclean without any optimizations as 
  detailed in Section~\ref{sec:optimization}. 
  \item \bcleannouc: This variant of \bclean does not employ UCs. 
  \item \bcleanII: This variant of \bclean incorporates \textbf{P}artition 
  \textbf{I}nference to our approach, which reduces unnecessary calculations by 
  partitioning the network. 
  \item \bcleanIII: This variant of \bclean enables both \textbf{P}artition 
  \textbf{I}nference and \textbf{P}runing based on sub-network semantics. 
  \item \pclean~\cite{DBLP:conf/aistats/LewASM21}: This method uses PPL to detect and 
  repair errors. \pclean requires the user to modify the relations and distributions 
  of the given datasets and outperforms other Bayesian data cleaning systems due to 
  its superior ability to encode expert domain knowledge.
  \item \rahabaran~\cite{mahdavi2019raha, DBLP:journals/pvldb/MahdaviA20}: This system 
  combines two methods, \raha and \baran, to construct a machine learning-based data 
  cleaning system. \raha is responsible for error detection using multiple error 
  detection mechanisms, while \baran repairs detected errors using various error 
  correction rules. We use the default settings, such as the number of labels, ML 
  models, and so forth.
  \item \holoclean~\cite{DBLP:journals/pvldb/RekatsinasCIR17}: This semi-supervised 
  inference system for data cleaning detects errors using signals and repairs them by 
  compiling integrity rules, matching dependencies, and statistical signals into 
  features in a Factor Graph. We run \holoclean with the signals either provided by the 
  dataset owners or collected by ourselves.
  \item \garf~\cite{Garf}: This rule-based method employs sequence generative 
  adversarial networks. It automatically learns rules from dirty data without any prior 
  knowledge and uses them to repair the data. We use the default settings as provided 
  in the paper.
\end{itemize}
Because DCs and labels used in \holoclean~\cite{DBLP:journals/pvldb/RekatsinasCIR17} and \rahabaran~\cite{mahdavi2019raha, DBLP:journals/pvldb/MahdaviA20} papers have not been 
released, we configure them by ourselves.

\noindent{}\textbf{Prior Knowledge.}
Table~\ref{DatasetSummary} also reports the statistical numbers of user inputs for all 
the competitors. For \bclean, we inject 12 regular expressions (detailed in 
Table~\ref{tab:ucs}), and all attributes in these datasets adhere to UCs. We avoid 
using specific values in regular expressions to prevent the leakage of ground truth. We 
assign data-quality experts to each dataset to formulate the UCs based on the attribute 
formats. Furthermore, these experts are also trained to label 20 tuples for \raha, 
correct 20 tuples as labels for \baran, craft DCs for \holoclean, and develop PPL 
programs for \pclean. Note that these baselines do not support expressions in UCs, and 
\bclean does not use inputs from them, such as DCs or PPL programs.

\noindent{}\textbf{Metrics.}
We use the following metrics to measure accuracy:
\begin{itemize} [leftmargin=*]
  \item \textbf{Precision}: The fraction of the correctly repaired errors over the total 
  number of modified cells.
  \item \textbf{Recall}: The fraction of the correctly repaired errors over the total 
  number of errors labeled with ground truth.
  \item \textbf{F1-score}: The harmonic mean of precision and recall.
\end{itemize}
To measure time and user cost, we collect the following metrics:
\begin{itemize} [leftmargin=*]
  \item \textbf{User time}: The total time spent by users to inject prior knowledge into 
  each baseline method. Initially, two experts are trained to understand the inputs and 
  outputs of each baseline, including DCs of \holoclean, UCs of \bclean, PPLs of \pclean, 
  and labeling of \rahabaran. User Time represents the average time these experts spend 
  to inject prior knowledge into a novel, unseen dataset. This metric assesses the 
  usability of the system for trained users. We exclude training time as it varies across 
  individuals.
  \item \textbf{Execution time}: The total runtime of the system modules to complete the 
  data cleaning task, given a dataset and the users' injected prior knowledge.
\end{itemize}

\noindent{}\textbf{Parameters.}
For \bclean, we set $\lambda=1$, $\beta=2$, and $\tau = 0.5$. These values reflect the 
principle that the lower the confidence in a tuple's relationship due to UC violations, 
the less trustworthy it is considered. A tuple with a confidence score lower than $\tau$ 
is deemed untrustworthy, with a penalty score of $\beta = 2$ applied to the relationship. 
Conversely, a credible relationship score is set to $\lambda = 1$. For other baselines, 
we apply the default settings in \pclean and \holoclean. We set labels of tuples to 
$20 + 20$ following the suggestions from the \raha and \baran papers.

\noindent{}\textbf{Environment.}
We conduct the experiments on a single machine equipped with 128GB RAM and 32 Intel(R) 
Core(TM) Xeon(R) 6326 CPU processors, each running at 2.90GHz. We run our experiments 5 
times and reported the average performance for both effectiveness and efficiency.

\subsection{Effectiveness and Efficiency}

\begin{table}
  \small
  \caption{Precision / Recall / F1-score on sampled \soccer}.
  \centering
  \vspace{-2ex}
  \resizebox{\linewidth}{!}{%
  \begin{tabular}{cccc}  
    \hline 
    \bclean & \holoclean & \pclean & \rahabaran  \\
    \hline
    0.345/0.931/0.504 & 0.919/0.551/0.689 & 0.15/0.665/0.244 & 0.523/0.133/0.212\\
    \hline
  \end{tabular}
  }
  \label{fig:holoclean-soccer}
\end{table}

\subsubsection{Data Cleaning Quality}
\label{sec:exp-cleanquality}
We assess \bclean's precision, recall, and F1-score against other competitors. As 
illustrated in Table~\ref{BaselineResult}, \bclean surpasses other techniques when 
the dataset has adequate relational information and a low error rate. \bclean 
achieves superior precision, recall, and F1-scores in both \hospital and \soccer 
datasets, which can be attributed to the high-quality BN generated by \bclean and 
its compensatory scoring model. For the \soccer dataset, \bclean significantly 
outperforms other techniques. The poor performance of \rahabaran can be attributed 
to the propagation of detection errors to the correction stage. \bclean also 
outperforms \garf, as the latter primarily focuses on generating explainable rules 
for error correction.

Due to the out-of-memory issue encountered with \holoclean, we randomly sample  
50,000 tuples from the \soccer dataset and evaluate the cleaning quality across 
all baselines. As shown in Table~\ref{fig:holoclean-soccer}, although \bclean's 
performance is inferior to \holoclean due to a limited relational context, its 
recall significantly exceeds that of \holoclean.

For the \flights dataset, despite its high error rate in comparison to the other 
datasets, \bclean maintains commendable performance, achieving results comparable 
to the best-performing method. 

Among the four versions of \bclean, the efficiency-optimized ones, \bcleanII and 
\bcleanIII, display similar performance to the non-optimized \bcleanI. These 
optimizations do not result in any significant impact on cleaning quality. 
Although \bcleannouc does not incorporate UCs, it remains competitive owing to the 
robustness of its BN construction and inference in the presence of noise, e.g., 
achieving an F1-score of 0.966 on \hospital, compared to 0.626, 0.730, and 0.715 
obtained by \holoclean, \rahabaran, and \garf, respectively.

Regarding other methods, \pclean achieves the best precision, recall, and F1-scores 
on the \flights dataset, however, this is largely reliant on precise domain 
knowledge modeling by experts. For the \soccer dataset, users find it challenging 
to articulate data distributions. Users can only estimate the distributions based 
on their observations and summaries, which results in less than satisfactory 
performance as evidenced by the results on \soccer.

\begin{figure}
    \centering
    \subfigure[Error distributions.]{\label{fig:error-dist}
        {\includegraphics[width=4cm,height=2.6cm]{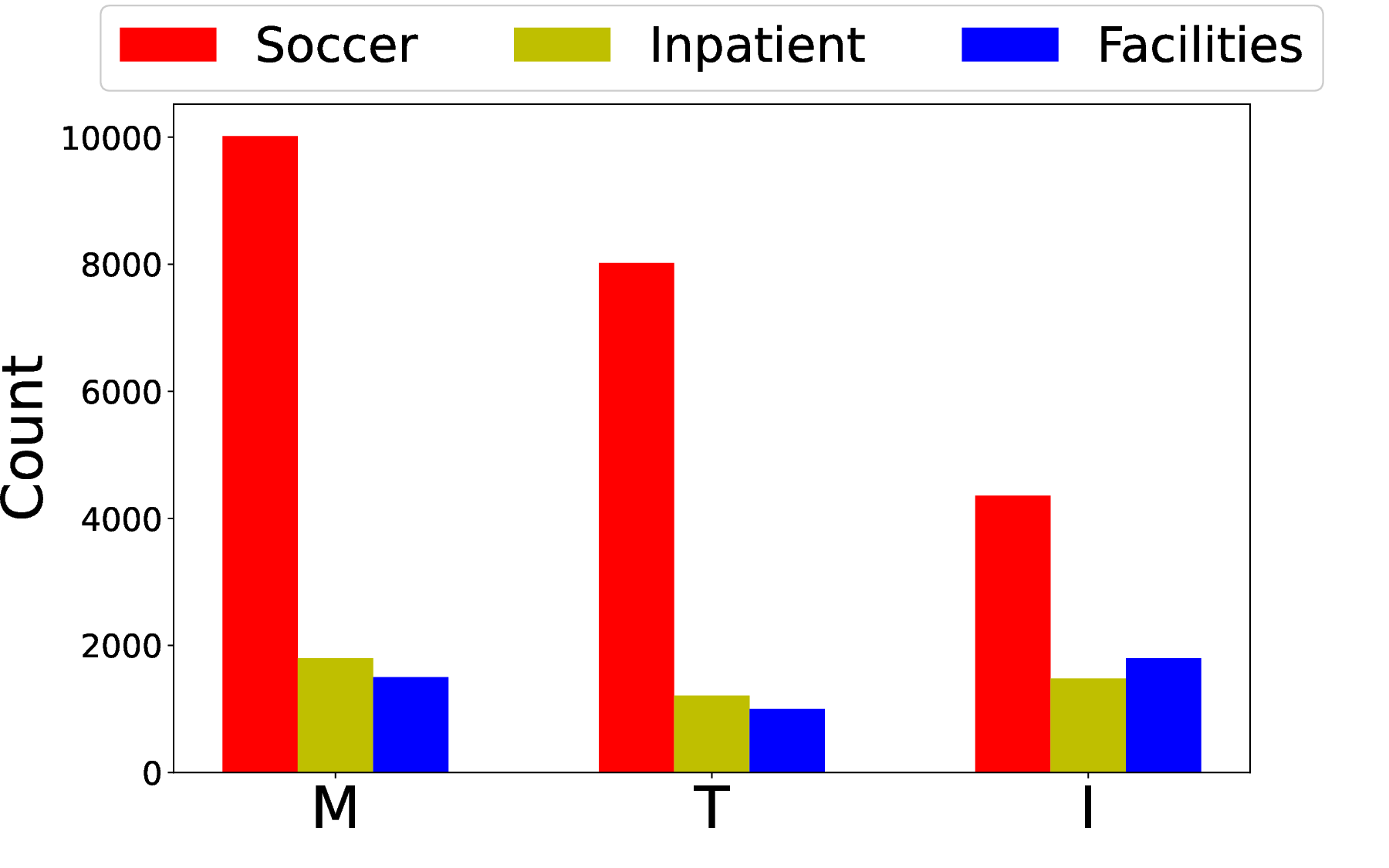}}}
    \hfill  
    \subfigure[\flights: varying error ratio.]{\label{fig:error-ratio-flight}
        {\includegraphics[width=4cm,height=2.6cm]{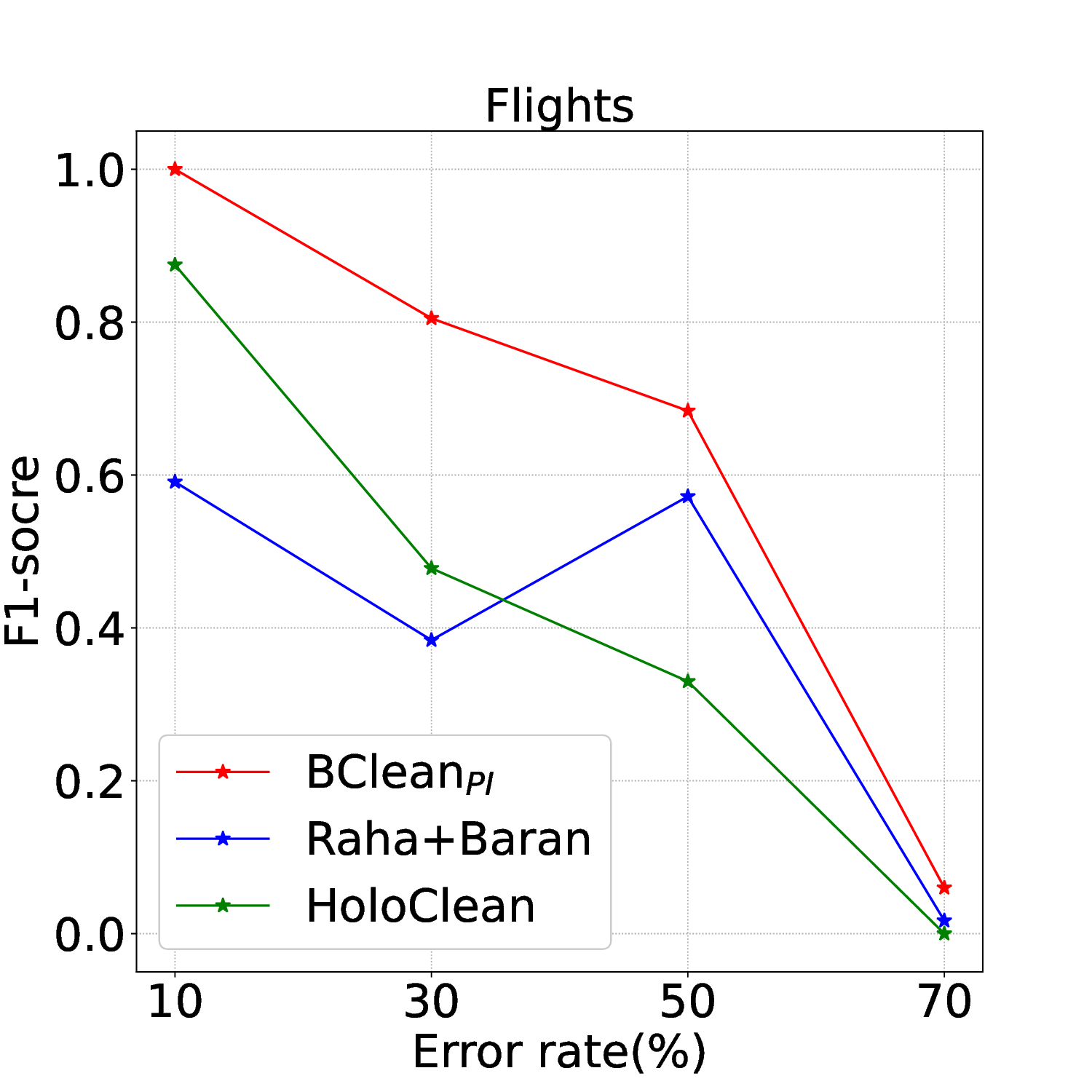}}}
    \hfill   
    \subfigure[\inpatient: varying error ratio.]{\label{fig:error-ratio-inpatient}
        {\includegraphics[width=4cm,height=2.6cm]{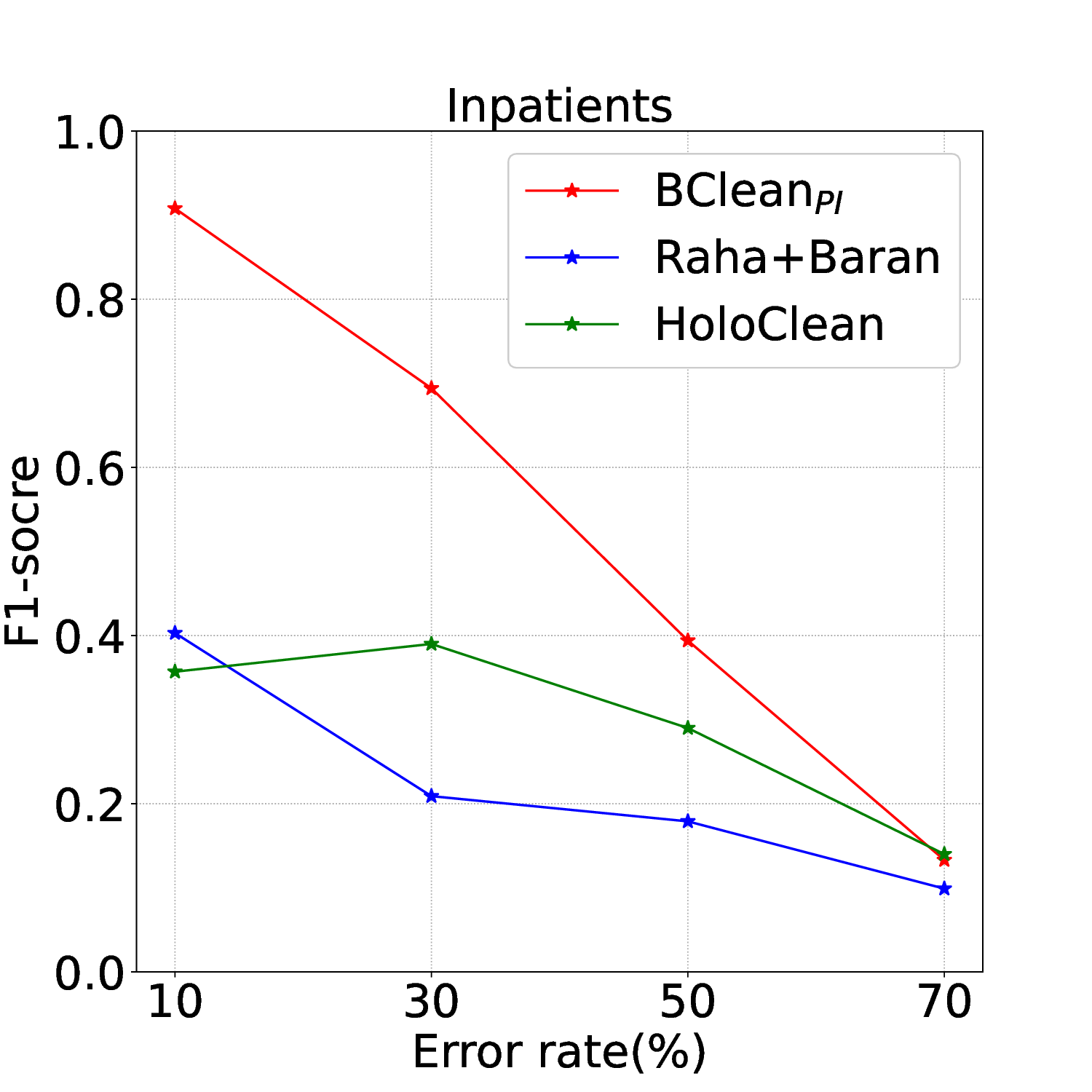}}}
    \hfill  
    \subfigure[\facilities: varying error ratio.]{\label{fig:error-ratio-facilities}
        {\includegraphics[width=4cm,height=2.6cm]{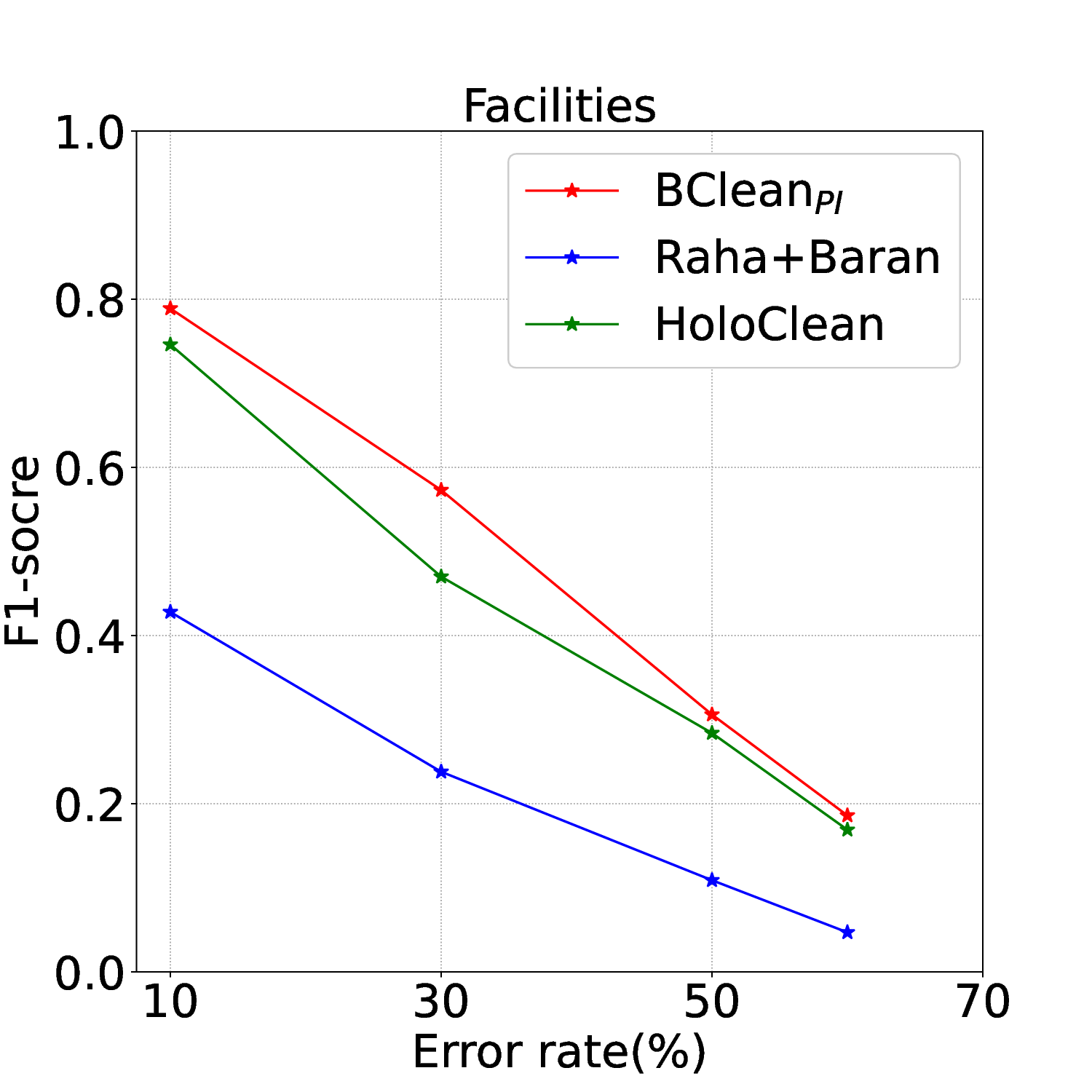}}}
    \hfill 
    \subfigure[\inpatient: swapping value errors.]{\label{fig:error-complex-inpatient}
        {\includegraphics[width=4cm,height=2.6cm]{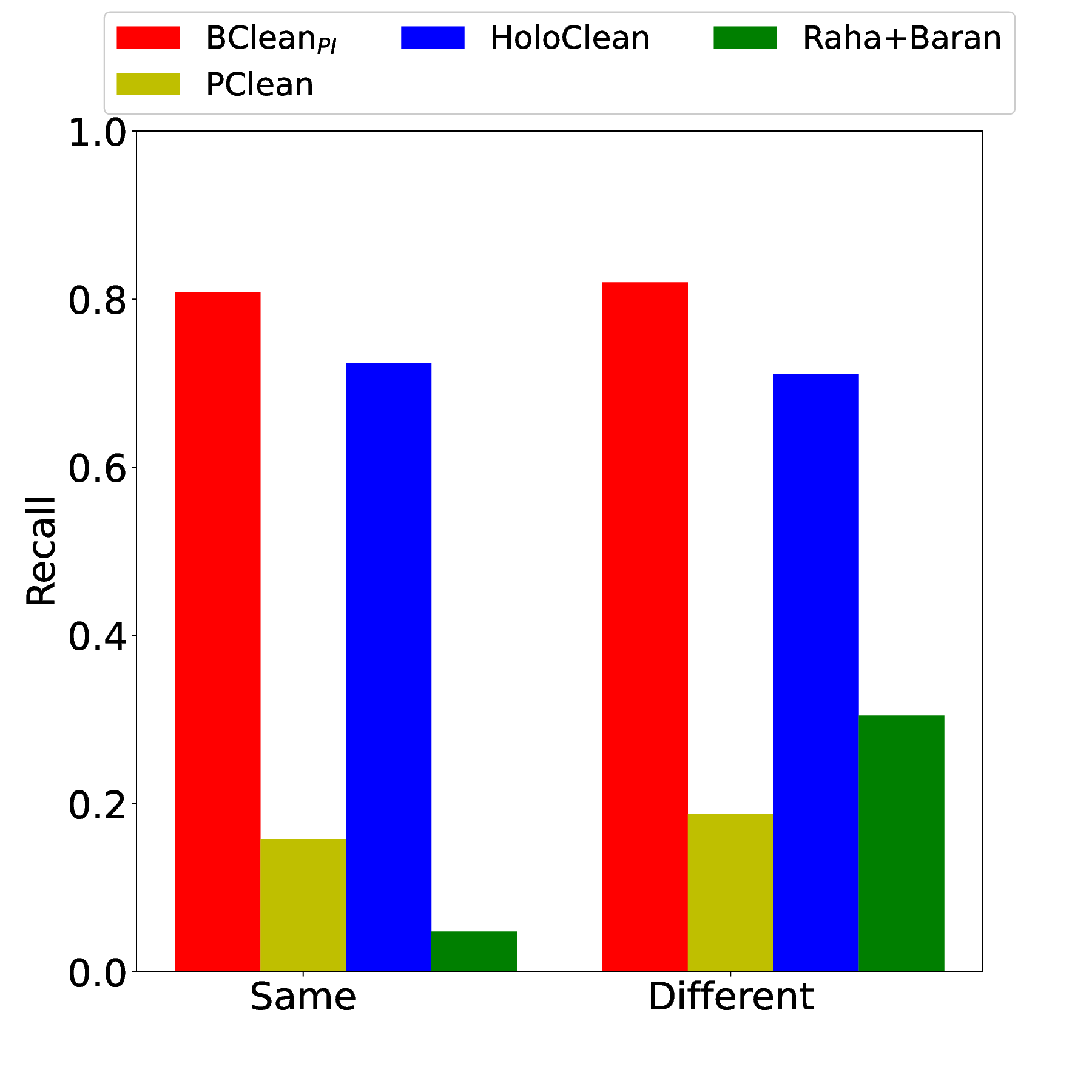}}}
    \hfill  
    \subfigure[\facilities: swapping value errors.]{\label{fig:error-complex-facilities}
        {\includegraphics[width=4cm,height=2.6cm]{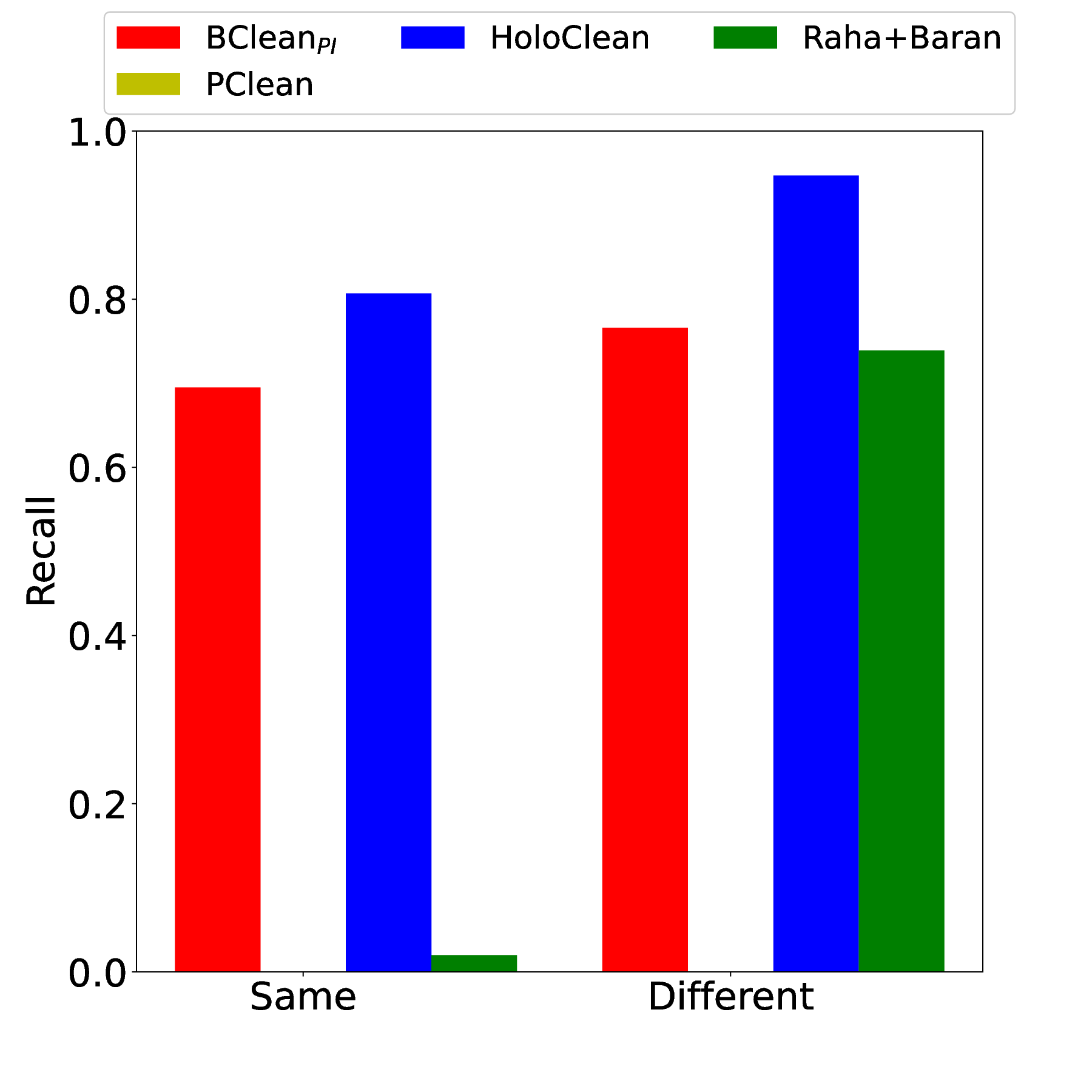}}}
    \hfill 
  \caption{Error analysis results.}
\end{figure}



\subsubsection{Error Analysis}
\label{sec:exp-error-analysis}
Next, we study how error types (T, M, and I, as specified in the error injection of 
the experiment setup) and ratio affect the performance. The distribution of errors 
across six datasets is displayed in Figure~\ref{fig:error-dist}. 

\etitle{Varying error types.} 
To assess the robustness of the competing methods, we evaluate \bclean's 
recall for each type of error within the \soccer, \inpatient, and 
\facilities datasets (Table~\ref{fig:recall-errors}). We do not report 
precision here, as it is challenging to determine which type of error a 
corrected value from the baselines originally belongs to. As shown in 
Table~\ref{fig:recall-errors}, \bclean consistently outperforms the other 
methods for each type of error, with higher recalls of 0.8, 0.9, and 0.9 
on average for T, M, and I, respectively, and up to 0.4, 0.5, and 0.5 higher 
than the best of the other methods. This showcases \bclean's robustness 
against various types of errors.

\begin{table}[t]
  \centering
  \small
  \caption{Recall for different types of errors (T, M, and I).}
  \vspace{-2ex}
  \resizebox{\linewidth}{!}{%
  \begin{threeparttable}
      \begin{tabular}{|l | c c c | c c c | c c c|}
      \hline
      \multirow{2}{*}{\textbf{Method}}  & \multicolumn{3}{c|}{\soccer} & \multicolumn{3}{c|}{\inpatient} & \multicolumn{3}{c|}{\facilities}\\
      \cline{2-10}
      \multirow{2}{*}{} & \textbf{T}  & \textbf{M} & \textbf{I} & \textbf{T}  & \textbf{M} & \textbf{I} & \textbf{T}  & \textbf{M} & \textbf{I} \\
      \hline
      \bcleanII & 0.997 & 1.000 & 0.990 & 0.840 & 1.000 & 0.843 & 0.683 &0.900   & 0.837 \\
      \pclean  & 1.000 & 0.568 & 0.953 & 0.323 & 0.760 & 0.477 & 0.0 & 0.0 & 0.0 \\
      \holoclean  & 0.749 & 1.000 & 0.923 & 0.954 & 0.612 & 0.949 & 0.804 & 1.000 & 0.851 \\
      \rahabaran  & 0.047 & 0.244 & 0.018 & 0.491 & 0.890 & 0.109 & 0.295 & 0.501 & 0.213 \\
      \hline
      \end{tabular}
  \end{threeparttable}
  }
  \label{fig:recall-errors}
\end{table}

\etitle{Varying error ratio.} 
We assess the competing methods by varying the error ratio from 10\% 
to 70\%, as depicted in Figure~\ref{fig:error-ratio-flight} -- 
\ref{fig:error-ratio-facilities}, following the evaluation in 
\cite{mahdavi2019raha}. The general trend reveals that all methods 
experience a decline in F1-score with increasing error ratios. However, 
\bclean demonstrates greater robustness and consistently outperforms 
the others. Even at a high error ratio of 70\%, \bclean maintains a 
relative performance advantage, with an F1-score across the three 
datasets on average 0.1 higher than the best-performing alternative.

\etitle{Swapping value errors.} 
We evaluate the performance of \bclean and other methods under 
conditions where 10\% and 5\% swapping value errors are randomly injected 
into the \inpatient and \facilities datasets. As illustrated in 
Figure~\ref{fig:error-complex-inpatient}, \bclean outperforms the others 
with a recall on average 0.1 higher than the best among the other methods. 
This validates \bclean's capability to handle swapping value errors errors, 
primarily due to the use of Bayesian inference featuring a BN and a 
compensatory scoring model.

\subsubsection{Runtime Analysis}
\label{sec:exp-runtime-analysis}
We report runtime statistics for all the competitors across the datasets. As 
demonstrated in Table~\ref{BaselineTime}, the user time required for \bclean 
is considerably less than that for all the competing methods, with the 
exception of \rahabaran. Due to its easy-to-deploy design, \bclean offers a 
more user-friendly learning curve and reduced complexity compared to the other 
methods.

Table~\ref{DatasetSummary} shows that \bclean users need to compose a few 
format-based UCs, focusing primarily on single attributes. \garf requires no 
user input because it is based on a generative model, while \rahabaran has the 
least user involvement among the competing methods.

Concerning execution time, \bclean takes slightly longer than \pclean but is 
faster than both \holoclean and \rahabaran on the \hospital and \flights 
datasets. However, when processing the larger \soccer dataset with 2M cells, the 
basic \bcleanI variant encounters efficiency issues, taking over 10 hours to 
complete. The optimized versions, \bcleanII and \bcleanIII, mitigate this problem 
by implementing partition inference and pruning. Their execution time is roughly 
on par with that of \pclean. When considering the total time cost, which includes 
both user and execution time, \bclean proves to be significantly more 
time-efficient than \pclean and \holoclean. Although \rahabaran boasts a shorter 
total time than \bclean, its lower quality reveals its limitations. It's also 
worth noting that \rahabaran requires significant time to infer correct values, 
as labeling tuples requires looking up values from other attributes and tuples, 
and necessitates user checks for contextual information. \garf, on the other 
hand, has a high execution time due to the generation of rules for 
interpretability.

\begin{table}[t]
  \centering
  \small
  \caption{Runtime of data cleaning methods, including user time (user) and execution time (exec).}
  \vspace{-2ex}
  \resizebox{\linewidth}{!}{%
  \begin{tabular}{|l|c|c c c c c c|}
    \hline
    \textbf{Method} & \textbf{Time} & \hospital & \flights & \soccer & \beers  & \inpatient & \facilities \\
    \hline
    \multirow{2}{*}{\pclean}  & user & $\geq$ 72h & $\geq$ 72h & $\geq$ 72h & $\geq$ 72h & $\geq$ 72h & $\geq$ 72h\\
    \cline{2-8}
    \multirow{2}{*}{} & exec & \textbf{16s} & \textbf{7s} & 30m 44s & 2m55s & \textbf{3m17s} & \textbf{1m32s}\\
    \hline
    \multirow{2}{*}{\holoclean}  & user & 15h & 12h & 14h & 15h & 15h & 15h\\
    \cline{2-8}
    \multirow{2}{*}{} & exec & 1m 40s & 36s & - & 1m37s & 4m14s & 6m2s\\
    \hline
    \multirow{2}{*}{\rahabaran} & user & 30m & 30m & 30m & 30m & 30m & 30m\\
    \cline{2-8}
    \multirow{2}{*}{} & exec & 1m 46s & 41s & \textbf{8m 59} & 3m2s & 10m36s & 10m55s\\
    \hline
    \multirow{2}{*}{\garf} & user & 0 & 0 & 0 & 0 & 0 & 0\\
    \cline{2-8}
    \multirow{2}{*}{} & exec & 5m24s & 1m57s & 18h30m & 2m8s & 26m48s & 30m10s\\
    \hline
    \multirow{2}{*}{\bclean} & user & 5h & 2h & 3h & 2h & 3h & 3h\\
    \cline{2-8}
    \multirow{2}{*}{} & exec & 25s & 17s & 10h 48m & 1m40s & 7h41m & $\geq$ 72h\\
    \hline
    \multirow{2}{*}{\bcleanII} & user & 5h & 2h & 3h & 2h & 3h & 3h\\
    \cline{2-8}
    \multirow{2}{*}{} & exec & 22 & 12s & 30m 42s & 31s & 7m57s & 17m16s\\
    \hline
    \multirow{2}{*}{\bcleanIII} & user & 5h & 2h & 3h & 2h & 3h & 3h\\
    \cline{2-8}
    \multirow{2}{*}{} & exec & 22s & 12s & 27m 46s & \textbf{30s} & 7m2s & 14m35s\\
    \hline
  \end{tabular}
  }
  \label{BaselineTime}
\end{table}

\subsection{Analysis of User Interactions}
\bclean involves user-interactions in two operations: defining UCs and 
manipulating the BN. We evaluate their impacts on the performance. 

\subsubsection{Impact of User Constraints}
We assess the significance of each UC by measuring the system's performance 
after the removal of some or all types of constraints. UCs consist of max 
length (\textsf{Max}), min length (\textsf{Min}), allowing null values 
(\textsf{Nul}), and patterns (\textsf{Pat}). We evaluate the system's 
performance with each constraint type removed, as well as after removing all 
constraints (\textsf{All}), and compare them to the complete version 
(\textsf{Com}).



\begin{figure}
    \centering
    \subfigure[Precision]{
        \includegraphics[width=0.45\linewidth]{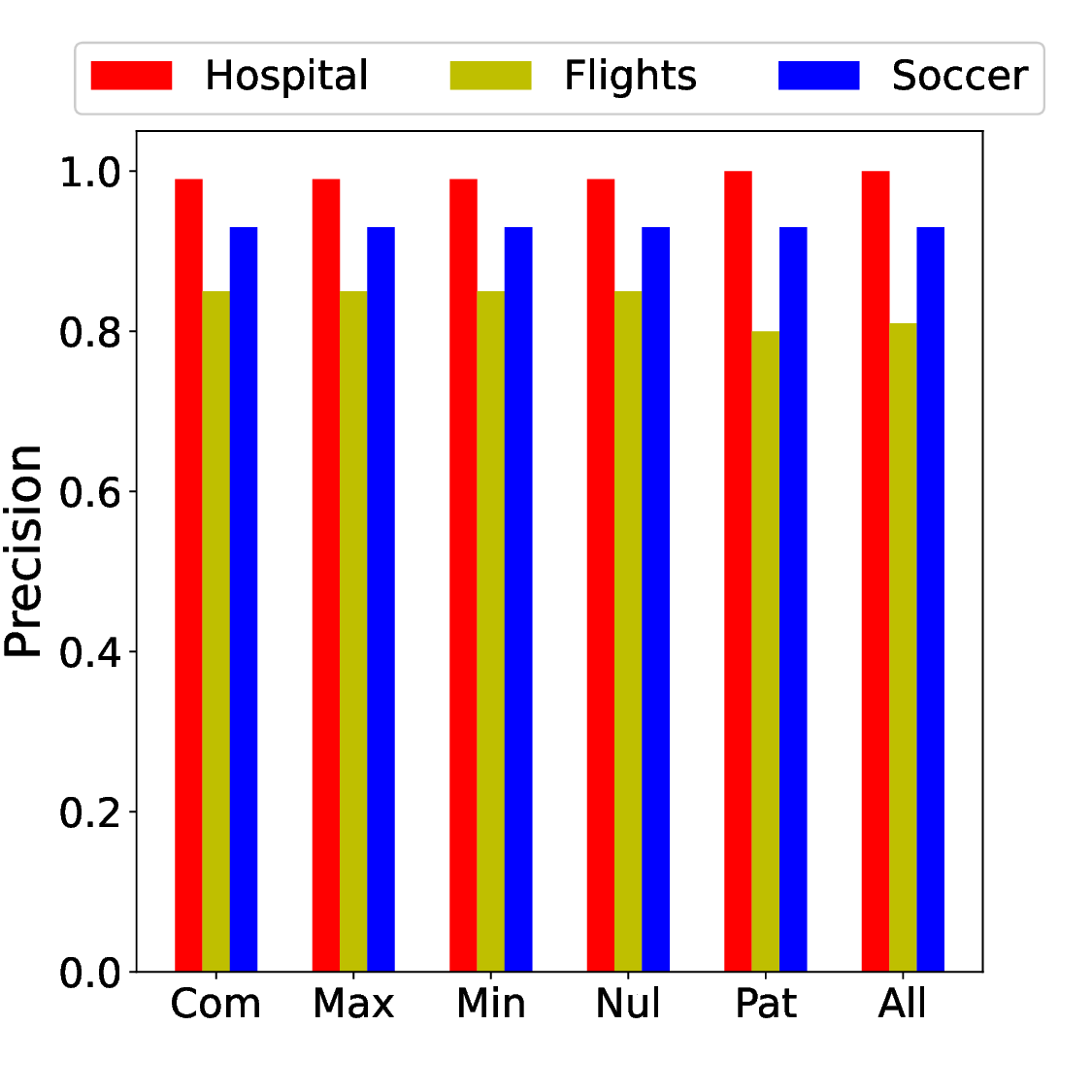}
        \label{fig:IncompleteConstraintPre}
    }
    \subfigure[Recall]{
        \includegraphics[width=0.45\linewidth]{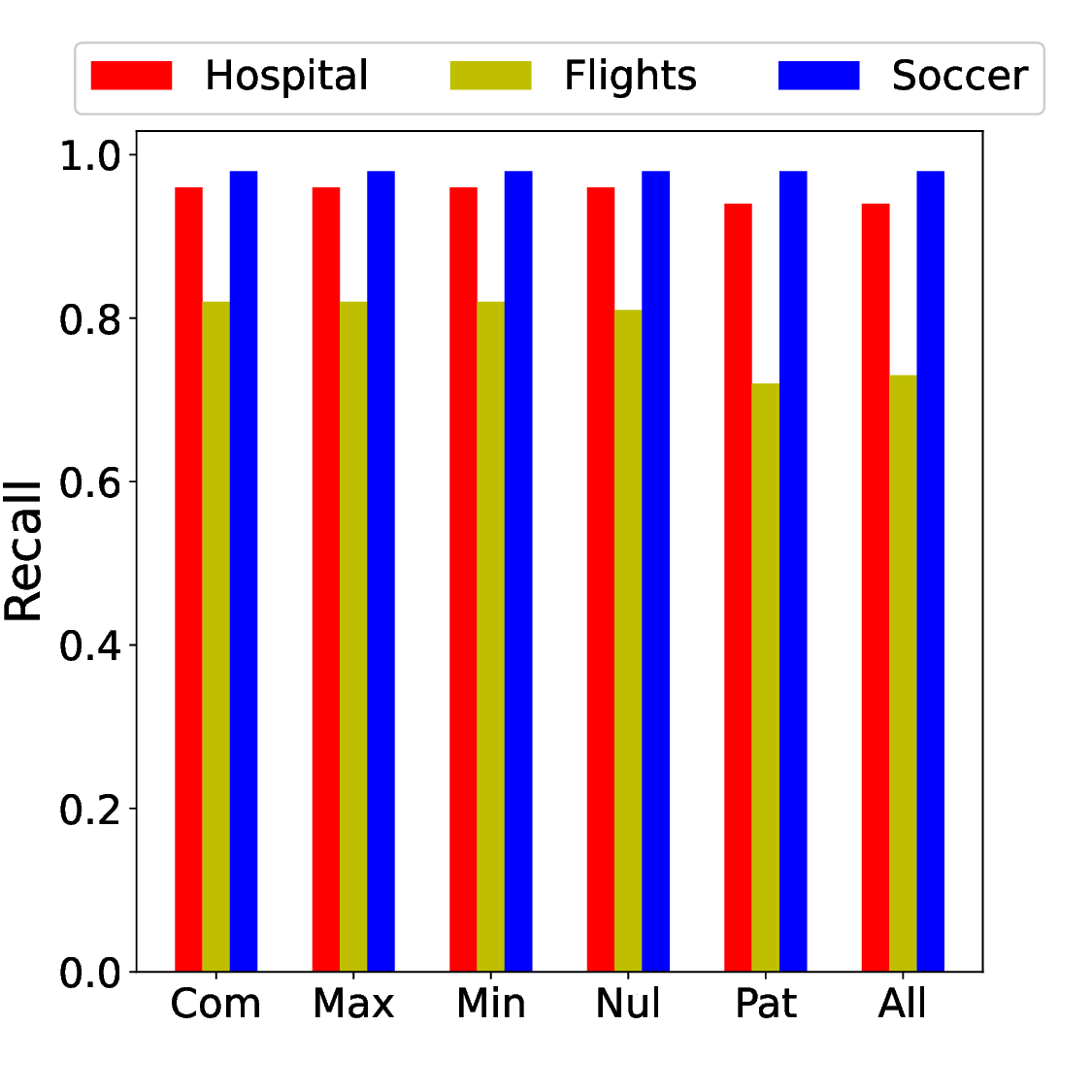}
        \label{fig:IncompleteConstraintRec}
    }
    \vspace{-2ex}
    \caption{Effect of incomplete UCs on precision and recall.} 
    \label{IncompleteUserInformation}
\end{figure}

As demonstrated in Figure~\ref{IncompleteUserInformation}, we find that the 
\textsf{Pat} constraint is the most influential. Changes in precision and 
recall are minor or even non-existent when \textsf{Max}, \textsf{Min}, and 
\textsf{Nul} are incomplete. When the user cannot provide the correct pattern 
(regular expression), or when no pattern exists to describe the dataset, a 
decline in precision and recall becomes apparent, as shown in 
Figure~\ref{IncompleteUserInformation} (a) and (b) respectively. For optimal 
performance, \bclean still needs expert input. Constraints provided by 
experts can directly exclude candidate values that do not meet the 
requirements before the inference process begins. This technique is extremely 
useful for data with many instances where the correct value is less than the 
error value and the error value does not meet the requirements, such as the 
\flights dataset.

In more detail, without pattern constraints, we receive two candidates with 
probabilities $Pr[g_{0} = ``7:10\ a.m.''] = 0.37$ and $Pr[g_{1} = ``7:21am'']=0.57$, 
and ground $g_{1}$ to be the final result in \flights. As shown in 
Figure~\ref{IncompleteUserInformation}, the system achieves sub-optimal 
inference and less effective error detection with lower precision (0.77) and 
recall (0.72) in the absence of a pattern. When a pattern is present, 
$Pr[g_{1}]$ is set to 0 prior to inference to prevent it from appearing in the 
final result. Thus, we only infer the ground truth repair within a restricted 
domain, yielding higher precision and recall. For \hospital, the pattern 
filters out the candidate value $g_{3}[\kw{ZipCode}] = $ ``1xx18'', as the ZipCode 
in \hospital must be a five-digit numeric string. This situation also occurs 
in \soccer.

In conclusion, while the absence of user constraints results in a slight 
decline in cleaning performance, the overall reduction remains within an 
acceptable range. This demonstrates that \bclean is user-friendly and suitable 
for users with limited expertise.

\subsubsection{Impact of Network Manipulation}
We conduct an experiment involving the modification of the BN through 
user interaction. The results indicate no difference between the states 
before and after the modification for the \hospital and \soccer datasets. 
After users' adjustment on BN, the \hospital dataset adds an edge 
${\kw{state} \rightarrow \kw{state\_avg}}$, resulting in one additional 
cell being cleaned, whereas the \soccer dataset displays no changes. 
However, for the \flight dataset, the BN structure obtained by the greedy 
search or FD-based method is incorrect, yielding cleaning results with a 
precision of only 0.217 and recall of 0.374. Following user adjustments, 
the results are improved to 0.852 precision and 0.816 recall, as shown in 
Table~\ref{BaselineResult}. In addition, such adjustments are with less 
than 5 minutes, suggesting that the user's effort is marginal.

\subsection{Parameter Tuning}
\label{sec:exp-parameter-tuning}
In \bclean, there are three parameters: $\lambda$, $\beta$, and $\tau$. To 
assess their impact, we fix two parameters at a 
time and vary the other. Tables~\ref{tab:HosptiallambdaVarrying} -- 
\ref{tab:HosptialtauVarrying} report the F1-score on the \hospital dataset. 
All three parameters have a minimal effect on the F1-score. 
This observation highlights that the performance of \bclean is stable and 
largely unaffected by parameter changes, aligning with the easy-to-deploy 
design of \bclean, which requires users to spend little effort on parameter 
tuning. 

\begin{table}[!t]
  \centering
  \small
  \caption{Varying $\lambda$ on \hospital, $\beta = 2$, $\tau = 0.5$.}
  \vspace{-2ex}
  \resizebox{\linewidth}{!}{%
  \begin{tabular}{|l|c c c c c c|}
    \hline
    \textbf{$\lambda$} & 0 & 1 & 2 & 5  & 10 & 15\\
    \hline
    F1  & 0.98096 & 0.98096 & 0.98096 & 0.98096 & 0.98096 & 0.98096 \\
    \hline
  \end{tabular}
  }
  \label{tab:HosptiallambdaVarrying}
\end{table}

\begin{table}[!t]
    \centering
    \small
    \caption{Varying $\beta$ on \hospital, $\lambda = 1$, $\tau = 0.5$.}
    \vspace{-2ex}
    \begin{tabular}{|l|c c c c c|}
        \hline
        \textbf{$\beta$} & 0 & 1 & 2 & 10  & 50\\
        \hline
        F1  & 0.97996 & 0.98096 & 0.98096 & 0.98096 & 0.98096\\
        \hline
    \end{tabular}
    \label{tab:HosptialbetaVarrying}
\end{table}

\begin{table}[!t]
    \centering
    \small
    \caption{Varying $\tau$ on \hospital, $\lambda = 1$, $\beta = 2$.}
    \vspace{-2ex}
    \begin{tabular}{|l|c c c c c|}
        \hline
        \textbf{$\tau$} & 0.1 & 0.3 & 0.5 & 0.7  & 0.9\\
        \hline
        F1  & 0.98096 & 0.98096 & 0.98096 & 0.97996 & 0.97996\\
        \hline
    \end{tabular}
    \label{tab:HosptialtauVarrying}
\end{table}

\section{Related Work}
\label{sec:related}

\noindent{\textbf{Data cleaning.}}
Data cleaning is an extensively-studied problem. We can broadly categorize 
the methods into five types. 

\begin{itemize}[leftmargin=*]
    \item \textbf{Generative methods}: Generative data cleaning methods 
    integrate error detection and correction components, often leveraging 
    probabilistic inference to iteratively clean datasets towards maximum likelihood~\cite{DBLP:journals/corr/abs-1204-3677, DBLP:journals/jdiq/DeHMCK16, DBLP:conf/icdt/SaIKRR19, doshi2003using}. Other notable methods in this category use 
    PPLs~\cite{DBLP:conf/aistats/LewASM21, milch20071} or user queries~\cite{DBLP:journals/pvldb/MahdaviA20, DBLP:conf/icdt/BertossiKL11}. Generative models are data-driven and do 
    not require any user labels, but many of them necessitate some prior 
    knowledge to achieve good performance.
    \item \textbf{Rule-based methods}: These employ data quality rules
    that check data inconsistency, such as FDs~\cite{DBLP:books/aw/AbiteboulHV95}, 
    conditional FDs~\cite{DBLP:journals/tods/FanGJK08, cfdclean}, 
    DCs~\cite{DBLP:conf/icde/ChuIP13}, user-defined 
    functions~\cite{DBLP:journals/pvldb/GeertsMPS13, DBLP:conf/sigmod/DallachiesaEEEIOT13}, 
    and regular expression entities~\cite{rees}. Rule discovery algorithms \cite{cfdsdiscovery1,cfdsdiscovery2,reesdiscovery} have been proposed to 
    mine these rules from relatively clean training data. Correspondingly,
    error detection~\cite{reesdetect} and data cleaning algorithms~\cite{Semandaq, DBLP:journals/pvldb/YakoutENOI11} have been designed to exploit these 
    rules to efficiently locate and correct errors in big data.
    \item \textbf{ML pipeline}: ML models have been widely used for data clean. 
    For instance, SCAREd~\cite{DBLP:conf/sigmod/YakoutBE13} adopts ML and 
    likelihood methods. In CleanML\cite{cleanml}, five error 
    types were evaluated, and the impact of data cleaning on the ML pipeline 
    was discussed. Picket~\cite{picket} employs a self-supervised strategy to 
    rectify data errors in ML pipelines. \cite{datarepairshapley} proposed a 
    novel framework based on Shapley values to interpret the results of any 
    data cleaning module. ActiveClean~\cite{DBLP:journals/pvldb/KrishnanWWFG16} 
    is a semi-supervised model with convex loss functions for data cleaning. 
    \item \textbf{ML imputation}: Additionally, many ML methods focus 
    on imputing missing values. A common approach is the denoising 
    autoencoder~\cite{dae}. More recent approaches use generative adversarial 
    networks (GANs) \cite{gain}, the attention mechanism~\cite{daema,aimnet}, or 
    both~\cite{cagain}. However, many of them require considerable effort in 
    tuning hyperparameters and training the model, especially for GAN-based 
    methods. 
    \item \textbf{Hybrid models}: These methods blend logical rules and ML 
    models so that rules are generated and used as features, and ML models are 
    applied for prediction using these features as inputs, e.g., 
    \cite{DBLP:journals/pvldb/RekatsinasCIR17, mahdavi2019raha, DBLP:journals/pvldb/MahdaviA20, heidari2019holodetect}. For instance, 
    \rahabaran~\cite{mahdavi2019raha, DBLP:journals/pvldb/MahdaviA20} applies 
    several feature engineering strategies (FDs, outliers, and patterns) to 
    generate features and then employs ML models for prediction. 
    HoloDetect~\cite{heidari2019holodetect} uses data augmentation and neural 
    networks to detect errors while minimizing human involvement.
\end{itemize}

In contrast to the aforementioned methods, \bclean employs a novel direction, 
executing approximate probabilistic inference with an easy-to-deploy 
configuration that includes UCs (e.g., regular expressions and statistical 
indicators), and a BN construction algorithm. User interactions are leveraged 
to fine-tune and retrieve the cleaning solution.

\noindent{\textbf{Bayesian network construction.}}
Existing BN structure learning methods include automatic greedy 
search~\cite{tzoumas2011lightweight, DBLP:journals/ker/Parsons11a, DBLP:journals/ml/TsamardinosBA06} 
and user-constructed networks~\cite{DBLP:conf/aistats/LewASM21, milch20071}. 
Automatic methods tend to be weakly robust to dirty data due to erroneous 
propagation, while user-constructed methods often incur high user costs when 
dealing with large datasets. In contrast, \bclean facilitates fine-tuning on 
structure learning via user interaction, which allows for minimal domain 
knowledge requirements while ensuring high data cleaning accuracy using the 
BN.

\noindent{\textbf{Bayesian inference.}}
Bayesian inference can be categorized into exact inference and approximate 
inference. Exact inference yields posterior probability distributions, but 
methods like variable elimination and belief 
propagation~\cite{DBLP:journals/ker/Parsons11a} can be computationally 
intensive and susceptible to erroneous propagation. Approximate inference, 
based on sampling techniques such as Gibbs 
sampling~\cite{DBLP:journals/pvldb/ZhangR14}, typically trades runtime 
improvement for accuracy. Exact inference is challenging to execute outside 
clean data, and approximate inference can propagate errors when sampling 
dirty data. \bclean generates approximate inference by 
partitioning the BN, thereby balancing accuracy with runtime efficiency.

\section{Conclusion}
\label{sec:concl}
We introduced \bclean, an easy-to-deploy Bayesian data 
cleaning system that requires minor user effort in specifying domain 
knowledge. \bclean automatically constructs a Bayesian network from an 
observed dataset and optionally adjusts the network through user 
interaction. User constraints, which cover a wide range of 
constraint forms such as dependency rules (e.g., functional dependencies 
and denial constraints) and arithmetic expressions, can be specified to 
reflect domain knowledge. \bclean detects erroneous data and infers 
correct values using Bayesian inference, which leverages the Bayesian 
network and a compensatory scoring model. To reduce the processing cost 
of data cleaning, we proposed a set of optimization techniques. Our 
experiments demonstrated the effectiveness and efficiency of \bclean 
and its superiority over alternative solutions.


\balance

\bibliographystyle{abbrv}
\bibliography{validate}



\end{document}